# Deep Learning-based Depth Estimation Methods from Monocular Image and Videos: A Comprehensive Survey

UCHITHA RAJAPAKSHA, FERDOUS SOHEL, HAMID LAGA, DEAN DIEPEVEEN, Murdoch University, Australia

MOHAMMED BENNAMOUN, The University of Western Australia, Australia

Estimating depth from single RGB images and videos is of widespread interest due to its applications in many areas, including autonomous driving, 3D reconstruction, digital entertainment, and robotics. More than 500 deep learning-based papers have been published in the past 10 years, which indicates the growing interest in the task. This paper presents a comprehensive survey of the existing deep learning-based methods, the challenges they address, and how they have evolved in their architecture and supervision methods. It provides a taxonomy for classifying the current work based on their input and output modalities, network architectures, and learning methods. It also discusses the major milestones in the history of monocular depth estimation, and different pipelines, datasets, and evaluation metrics used in existing methods.

## 1 INTRODUCTION

Depth estimation aims to recover the 3D geometry of a scene from one or multiple images. It is fundamental to many applications, including autonomous driving, 3D reconstruction, augmented reality, 3D digital entertainment, robotics and biology. Monocular depth estimation, i.e., depth estimation from a single RGB image or video, is popular due to its practicality for in-the-wild and low cost. It is, however, a very challenging task due to the loss of information in the 3D to 2D projection process. While humans can sense the depth of images using monocular cues and prior knowledge, this is an ill-posed problem for computers.

Existing depth estimation methods rely on active sensors (e.g., laser scanners) or passive sensors such as RGB cameras. Although active sensors such as Light Detection Range (LiDAR) and Time of Flight (ToF), which transmits energy (light, sound, or radio waves) onto surfaces, directly provide accurate 3D information, they are often expensive and time-consuming. Also, obtaining depth for small and close objects in short ranges using active sensors is challenging, as the sensitivity depends on the effective range of the sensors. Examples of image-based (passive) depth estimation methods include structure-from-motion, shape-from-X where X is a silhouette, texture, shading, or (multi-view) stereo [43]. Most methods require two or more images of the same scene and accurate

Authors' addresses: Uchitha Rajapaksha, Ferdous Sohel, Hamid Laga, Dean Diepeveen, Murdoch University, 90 South Street, Murdoch, WA 6150, Australia, uchitha17669@gmail.com,f.sohel@murdoch.edu.au,h.laga@murdoch.edu.au,d.diepeveen@murdoch.edu.au; Mohammed Bennamoun, The University of Western Australia, 35 Stirling Highway, Crawley, WA, 6009, Australia, mohammed.bennamoun@uwa.edu.au.

calibration of the cameras. Such methods are not always practical for in-the-wild applications such as autonomous driving and augmented reality. The interest in monocular depth estimation methods has rapidly increased since they are less expensive, less time-consuming, and do not require complex camera calibration.

Today, many researchers formulate the monocular depth estimation problem as a machine learning task. Recent developments in deep learning also fuel the increasing interest in deep learning-based methods for monocular depth estimation. This paper aims to provide a comprehensive survey and structured review of the latest deep learning-based methods on monocular depth estimation. We present a broad taxonomy of the existing methods considering more than 160 key papers appearing in leading conferences and journals between January 2012 and January 2024. The following list summarizes the main contributions of this paper.

- We highlight the importance of a broad taxonomy and a comprehensive survey on monocular depth estimation. We review more than 160 papers published between 2012 and 2024 in major computer vision, computer graphics, and machine learning journals and conferences.
  We provide a taxonomy of existing methods that consolidate their input and output modalities, network architectures, degree of supervision, types of datasets and domain adaptation methods used to train and evaluate deep learning models.
- We present several aspects of challenges in the field. We first describe the typical factors that make single-image or video-based depth estimation challenging as general challenges. We highlight the challenges in modeling the task as a deep-learning problem (Section 3).
- We present a generic pipeline and a baseline architecture to compare the literature based on the network architectures they adopt at each step in the reconstruction pipeline. We also compare methods based on the degree of supervision and loss functions used to train existing models (Sections 4, 5 6 and 7).
- We review commonly used datasets for training and testing deep learning models and present a summary of more than 20 datasets (Section 9). We also discuss the standard evaluation metrics in Section 10.
- We discuss future research directions based on the identified gaps (Section 11).

This paper complements other survey papers on deep learning-based 3D object reconstruction [49] and stereo-based depth estimation [77, 119] by focusing only on depth estimation from monocular image and videos. In addition, much research has been done on 360° monocular depth estimation [130]. However, we have left this topic out of the scope of our survey, as we think 360° is so large that it can be a topic of another survey paper.

## 2 BEYOND EXISTING SURVEY ARTICLES

Although several survey papers [6, 9, 14, 62, 107, 150, 156, 170, 170, 202] on monocular depth estimation are available, they provide a limited scope. Most importantly, they lack recent advancements in this context. Table 3 (Section D in Appendix A) summarizes existing surveys. We provide a short summary of the limitations of them in Section E of Appendix A. In contrast, ours is the first to survey the seven aspects: challenges, input and output, network architectures, degree of supervision, datasets, evaluation metrics, and future directions in monocular depth estimation. We include more than 160 research papers published between 2012 and 2024, significantly covering more papers than existing surveys. Compared to existing surveys, we present a broad taxonomy that covers input and output modes, network architectures, and supervision modes. Our taxonomy covers a variety of research with the most recent and advanced architectures, indoor and outdoor depth estimation methods, and domain adaptation methods. To our knowledge, we are the first to use CNN as the baseline auto-encoder architecture and compare existing work based on the baseline. We have surveyed degrees of supervision; 3D supervised, self-supervised, semi-supervised, weakly supervised, and online learning. Finally, we provide a detailed discussion on future directions in this field based on the gaps identified after the survey.

## 3 PROBLEM STATEMENT, CHALLENGES AND TAXONOMY

Given a single image $I$, monocular depth estimation aims to learn a function $f$ that estimates the per-pixel scene depth $d$ where,

$$d = f(I). \tag{1}$$

Existing deep learning-based methods for monocular depth estimation focus on designing deep neural architectures that approximate the function $f$, and efficiently train them to achieve the best performance. This is not a straightforward task, as they should also address the following challenges.

### 3.1 Challenges

Here, we discuss the challenges based on whether they are related to the general monocular depth estimation problem or are specific to the deep learning-based approaches.

*3.1.1 General challenges.* Estimating depth from a single RGB image is a challenging task due to several reasons:

- **3D-to-2D mapping ambiguity**: Depth ambiguity is one of the main challenges when recovering 3D information from 2D images. When an RGB camera captures an object in the 3D space, it loses the 3D properties, such as pose, volume, and 3D shape. A 2D image only contains the texture and 2D projection of the object in the form of a 2D pixel grid. As a result, the 2D representation of an object can have multiple 3D interpretations, which result in depth ambiguity. Unlike stereo and multi-view images, which are rich in stereo cues, monocular cues alone cannot resolve such depth ambiguities. Consequently, obtaining reliable depth results using a single image is difficult.
- **Camera parameters estimation**: Recovering the absolute depth requires estimating the camera's intrinsic and extrinsic parameters. Intrinsic parameters such as focal length and aspect ratio decide the mapping of each point in camera coordinates to the pixel coordinates. When accurate camera parameters are available, depth estimation methods use them to obtain even a rough depth estimation such that the estimated depth renders a similar image as the input image. Therefore, camera poses enable comparing the actual and estimated depth using single images. However, in practical scenarios (e.g., in-the-wild reconstruction), camera parameters are unavailable, making the process challenging.
- **Brightness inconsistency in scenes**: Depth estimation from monocular images is difficult due to the color and illumination variability in different inputs [40, 189]. Due to brightness inconsistencies, two images representing the same depth map can have different color distributions. Due to such variations, for example, inconsistencies during day and night, multiple inputs of the same scene under different weather conditions produce inconsistent depth distributions [13]. Consequently, deriving a direct relationship between the color of a pixel and the depth is challenging.

Using videos instead of still images helps address some of the aforementioned challenges. Videos, however, bring another set of challenges, such as:

- **Temporal consistency and smoothness**: Maintaining depth consistency of the outputs across the frames is essential. The presence of inconsistent depth values in a sequence of video frames causes unwanted flickering effects in the depth output [99, 151]. Avoiding such flickering effects is critical for real-world applications. For instance, humans are sensitive to depth inaccuracies, as such inaccuracies cause discomfort to the eye with a small change or inconsistency in the output sequence. Therefore, maintaining temporal consistency and smoothness of the estimated depth outputs is essential yet challenging.
- **Moving objects and moving camera**: Monocular videos often contain objects that move rigidly and non-rigidly. The latter makes matching pixels across images very challenging. When a moving camera captures input videos, depth estimation methods should first disentangle the camera motion from objects' motion to ensure accurate depth estimation. Therefore, inaccuracies in camera motion estimation also produce inaccurate depth estimations [65]. Some approaches assume static environments and rigid movements to reduce the impact of moving objects. Such assumptions are not realistic for real-world applications.

In addition to the general challenges, we highlight the major challenges related to deep learning below.

*3.1.2 Challenges of deep learning-based monocular depth estimation.* Using prior knowledge, humans easily understand close and far away objects in single RGB images. Inspired by this, deep learning-based methods formulate the monocular depth estimation as a learning problem. The following challenges are specific to them;

- **Training-related issues**: Deep learning methods learn the prior knowledge from the data presented to the models at the training stage. However, acquiring sufficient training data and producing accurate results for unseen data is challenging. The following constraints are some of the training-related issues.
  - **Acquisition of accurate 3D ground-truth**: Learning-based methods require accurate 3D ground-truth depth for training. Acquiring such ground-truth is tedious, time-consuming, and requires expensive sensor setups. The available ground-truth depth is sparse (partial and incomplete) as well. Sometimes correct ground-truth depth is not paired with the 2D images in their corresponding resolution [43].
  - **Generalizing and adapting to new domains**: One of the major limitations seen in many approaches is that the learned depth strongly relies on the training data. Therefore, the models usually show low accuracy on previously unseen images. Although these methods perform well for their task-specific domain, they perform poorly on new domains [39, 154]. Moreover, training only on a limited dataset leads to the domain overfitting problem. Therefore, developing general approaches for different domains (e.g., day, night, indoor and outdoor) is a challenge [93]. Addressing domain shift-related issues (e.g., shifting from daytime to nighttime images) is essential for real-time applications to ensure the models perform well under raw feed with minimum pre-processing.
- **Computation time and resources**: Deep learning-based methods are expensive regarding memory requirements and computation time. They often require GPUs during both training and runtime. In practical scenarios, achieving high accuracy on resource-constrained devices such as mobile devices, embedded systems, and virtual and augmented reality devices remains a challenge.

While deep learning can be challenging, such methods record the best results in monocular depth estimation.

## 3.2 Taxonomy

Figure 1 provides the detailed taxonomy of the existing methods based on their input modes during training and runtime, network architectures, degree of supervision, datasets and domain adaptation.

*3.2.1 Input and output.* This survey focuses on monocular depth estimation methods where the input during inference is either a single RGB image [39, 43, 62, 63, 76, 78, 113, 116, 189, 194] or a monocular RGB video [112, 162, 165]. For most methods, the input during training is single RGB images annotated with either dense or sparse ground-truth depth values [63, 78, 134]. Other methods require stereo images or monocular video sequences to enable self-supervision without ground-truth depth annotation. Some use monocular videos [127, 208] to train the models. Temporal and geometric consistencies of the scene throughout the video are useful in producing depth alignments and estimating depth even without ground-truth depth. Other approaches use stereo input for training [39, 43, 76, 113, 116, 189, 194]. Stereo input can be either stereo image pairs [21, 39], stereo video stream data, or multi-view stereo. At least two rectified stereo images, i.e., left and right views, are required to obtain the disparity cues present in the scene and thereby acquire depth information during training. Although such methods require stereo images during training, they only require monocular images during inference. Hence, they come under monocular depth estimation.

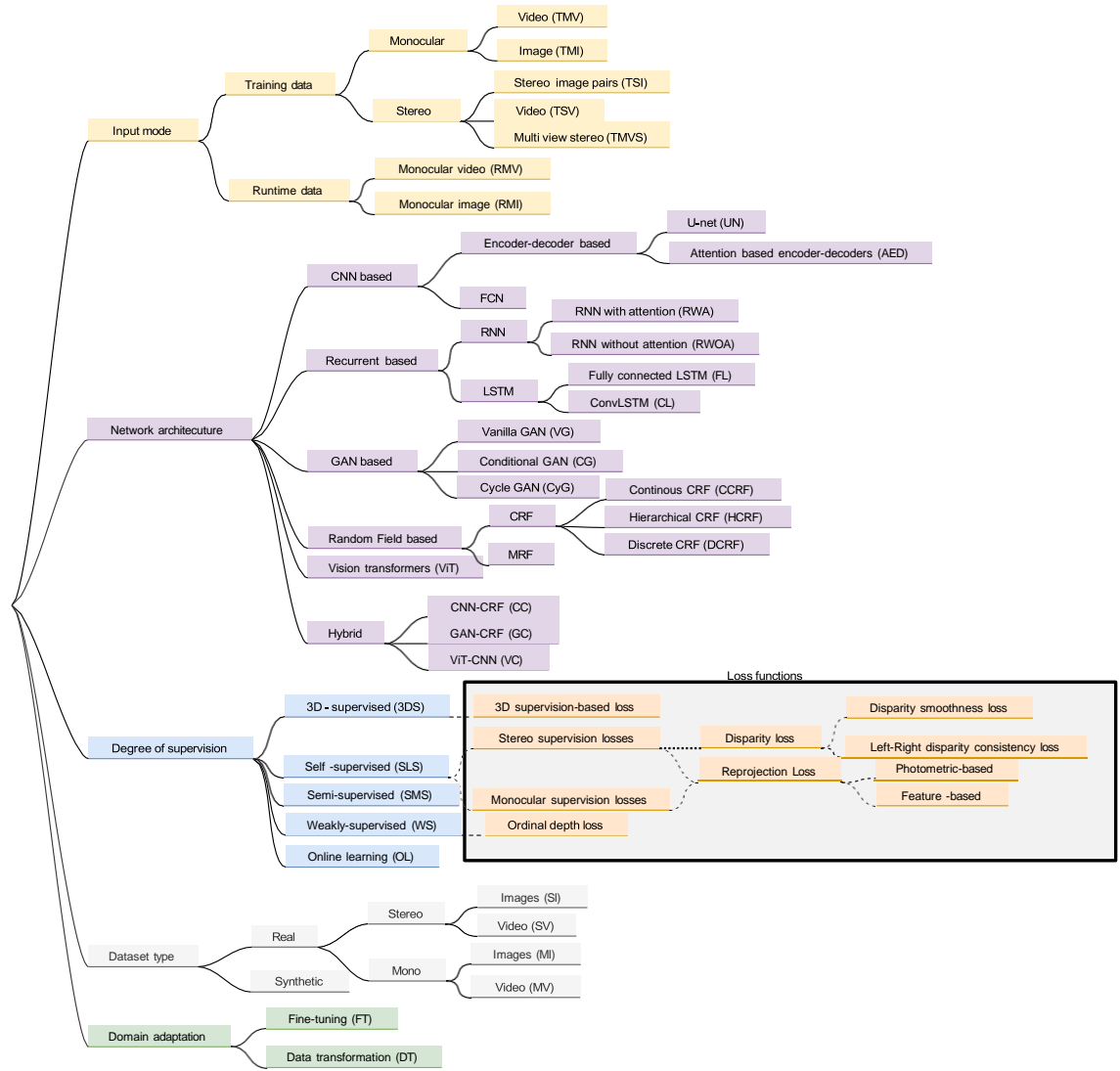

Fig. 1. Taxonomy of deep learning-based monocular depth estimation techniques. The gray color box shows the summary of existing loss functions based on the type of supervision.

In terms of output, existing methods produce dense or sparse depth maps [39, 43, 62, 63, 63, 76, 78, 78, 113, 134, 189]. A depth map is an image indicating depth values at each pixel of the input. Early methods [121] produce low-resolution depth maps.

Most methods compute the relative depth with respect to the objects present in the scene. Others attempt to estimate the absolute (metric) depth [88]. Some methods also estimate uncertainty maps, indicating how confident the network is in its depth estimation task at each image pixel [186]. They estimate the camera parameters and an object-level segmentation of the scenes [51]. Some approaches produce a probability distribution of plausible depth for a given input image [140, 168].

*3.2.2 Pipeline and network architectures.* We present an extended pipeline with **(i)** masking and data augmentation, **(ii)** feature learning, **(iii)** feature selection, **(iv)** coarse depth estimation, **(v)** fine depth estimation and **(vi)** post-processing. However, the network architectures in existing methods differ mainly based on the type of networks they use at coarse depth estimation and refinement stages. Section 6 discusses them in detail.

*3.2.3 Degree of 3D supervision.* We classify existing training procedures based on the degree of 3D supervision they require. For instance, some methods require images paired with their ground-truth dense depth maps [83, 134, 172]. Such 3D supervised methods are expensive. Weakly and semi-supervised methods address this issue by using either sparse 3D annotations or ordinal 3D annotations, which are easier to acquire [3, 46, 76, 184]. Self-supervised methods, also known in the literature as unsupervised methods, only require 2D images and videos for training [43, 189, 194].

*3.2.4 Training datasets.* We provide a summary of the datasets and classify them based on the type of data as real and synthetic and further divide them as monocular and stereo.

*3.2.5 Domain adaptation.* We also classify the domain adaptation methods used during training: fine-tuning and data transformation.

## 4 MONOCULAR DEPTH ESTIMATION PIPELINE

Figure 2 provides an overview of the general pipeline that most existing methods for depth estimation from single images and videos follow. Early methods used a pipeline composed of four main stages: feature extraction, coarse depth estimation, fine depth estimation and post-processing. Recent methods achieve better accuracy by incorporating different strategies for data augmentation with masking and feature selection.

### 4.1 Masking

Humans easily recognize the monocular cues to understand the relative depth of objects in a given image. Examples of cues include the horizon, vanishing point, relative object size, light and shade, interposition and texture gradient. Therefore, humans do not require all the information provided in the entire image to understand the depth, as some small regions of the visual field associated with the cues are enough for the task. Similarly, recent deep learning models leverage the same idea to learn depth from a sparse set of data in the image instead of the whole image [34]. As an initial step, masking aims to filter a sparse set of data required for training from the input images. Therefore, we can formulate the monocular depth estimation problem as a function of input $I$ , a mask $M$ and a masking operation⊗ [54].

$$d = f (I \otimes M). \qquad (2)$$

This mask can be a simple binary mask, which determines the valid or invalid pixels for the sparse input, or a continuous mask quantifying the significance of each pixel for the final outputs as a weight or probability. Continuous masks perform better than binary masks [100].

Existing methods formulate the problem of identifying the relevant pixels as an optimization problem. They optimize the model to identify the smallest number of image pixels from which the deep learning models, i.e., CNNs, can estimate a depth map. Therefore, the estimated depth map should be as close as possible to that of the estimation produced from all pixels in the image [37, 54]. Hu et al. [54] show that deep learning models can infer depth using a sparse set of regions in an image, contributing more to depth estimation. Such regions depend on the input domain type. For instance, the region around the vanishing point is vital in learning to estimate depth for an outdoor scene. Therefore, masking the outdoor dataset by selecting the pixels around the vanishing point produces a sparse dataset, enabling higher training efficiency. In a similar attempt, Dijk et al. [34] show that models rely on the location (vertical position) of objects in the image rather than the object's appearance or size. Apart from the aforementioned cues captured by human vision, there are depth-specific structural features to augment the data for depth estimation frameworks. For example, S2R-DepthNet [25] extracts a general domain-invariant structure map from the inputs to obtain a structural representation. Thus, the structural representation ignores irrelevant data and encodes only the important structural features.

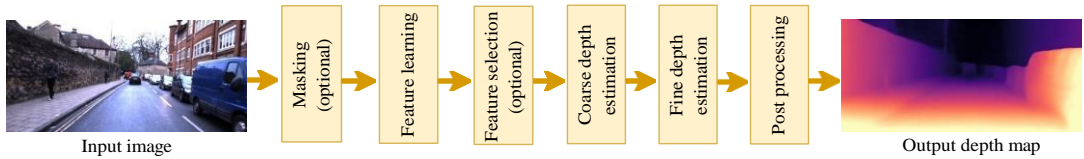

Fig. 2. The general pipeline for depth estimation from monocular images. The pipeline consists of six stages: masking, feature learning, feature selection, coarse depth estimation, fine depth estimation and post-processing.

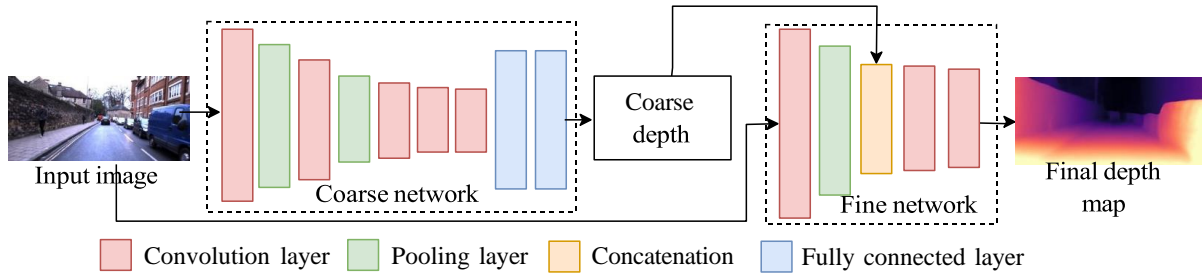

Fig. 3. Coarse-to-fine refinement with global and local feature fusion [36]. This example also covers the coarse and fine depth estimation stages of the general pipeline.

Masking strategies enable the models to learn depth with cues rather than just memorizing. Masking helps to develop robust deep learning models that can estimate depth accurately, even for out-of-distribution samples.

## 4.2 Feature Learning

Deep learning models learn features from the sparse or dense data provided as input. The features include low-level features such as color, intensity, texture and lighting, and high-level features such as pixel-level context, image-level context, occlusion boundaries, perspective and semantic similarity. The process involves encoding the input image using a backbone network (see Section 6.1). A generic feature extractor network performs convolutions to produce feature maps at different scales. Extracting low-resolution features, which capture long-range context, and fine-grained features describing local context leads to better results in depth output [148].

Some methods divide the input image into image patches before estimating depth. The resolution of the input images or patch size, the receptive field of the network, and the kernel size of the convolutional filters of its architecture determine the model's capability to learn the features. Since the accuracy of the estimated depth values improves with input image resolution, some methods extract features at different input resolutions [52]. Some use a patch-based feature extraction rather than encoding the entire image to capture more fine-grain details in the local neighbourhood [106]. Several works use dilated convolutions to extract context information [123, 133, 176].

Dilated convolutions allow the models to expand the field of vision in learning features by retaining the spatial resolution of the feature maps.

## 4.3 Feature selection

The efficiency of the deep learning models depends highly on the features they learn. Although early methods treated all learned features equally, recent methods show higher accuracy with attention to the best features. For instance, fine details account for the quality of the output depth map produced by the models [50]. Features that describe fine details help to preserve more structural details in the final output. Therefore, selecting such features from the extracted features and learning to pay attention accordingly is crucial. Section 6.3 discusses more details on feature selection methods using attention mechanisms.

## 4.4 Coarse-to-fine depth estimation

Estimating depth from the selected features involves coarse depth estimation and fine depth estimation.

*4.4.1 Coarse depth estimation.* Progressively refining the prediction from previous layers to obtain a fine output is common in deep learning methods. Similarly, monocular depth estimation commonly produces a coarse depth as an initial step. Eigen et al. [36] propose to train two networks in a coarse to fine manner. The first network focuses on the global scale of the input image to produce a global depth estimation. Coarse estimation aims to output a depth map indicating only the overall structure of the scene. Fully connected layers are often used in the decoder of the coarse network to capture the entire image in the field of view, while the middle and lower layers of the encoder combine information from the different regions of the image. For this purpose, the coarse network focuses on several global features, such as the vanishing point, object locations, and alignment. Max pooling operations integrate global features. The fine network takes the output of the coarse network for further refinement with local information.

*4.4.2 Fine depth estimation.* Two ways to refine the coarse depth are (i) global-to-local and (ii) semantic segmentation-based depth fusion.

· **Global-to-local**: Global-to-local feature fusion is a popular coarse-to-fine procedure in CNN architectures (see Figure 3). Finer details captured at a local feature level yield higher accuracy for depth estimation. The refinement network takes the input image to produce a feature map with local features and concatenates it with the coarse depth feature map during encoding. Zero padding reduces any spatial size reduction during convolution layers throughout local refinement to obtain final output in a relatively high resolution. Several other studies [16, 27, 39, 66, 76, 78, 199] also proposed architectures that use coarse-to-fine strategy.
· **Semantic segmentation-based**: Semantic segmentation-based monocular depth estimation utilizes the depth consistency within semantic regions of the input image. They segment a given image into sub-regions based on semantic similarity. Such methods isolate objects within an image from their background and consider the sky or ground as separate segments. Wang et al. [160] propose to predict the depth for each segment as a normalized depth map which takes values in the range of zero and one. They then aggregate the produced normalized depth for each segment based on global context in a divide-and-conquer manner. However, their method requires segmentation annotations for training, which constrains the training process only to the datasets that contain semantic labels.

Semantic segmentation shows better results in depth refinement because it produces consistent depth structures in the output compared to those without semantic features.

### 4.5 Post-processing

Post-processing steps improve the quality of the estimated depth maps. Properties of the objects, such as shadows and occlusions in the input scene, are likely to create artifacts in the estimated depth map. Post-processing steps overcome such errors during both training and runtime. Godard et al. [43] propose a simple yet efficient post-processing method for estimating the uncertainty of the output depth map. They propose to estimate depth twice for the same input, once for the original input and again for the flipped input image. The mean value of the depth values of corresponding pixels in both depth maps is the final depth of each pixel. Following this, many methods use horizontal and vertical flipping as an efficient post-processing step in their models [120, 129]. Peng et al. [113] also propose a self-distillation-based selective post-processing step to adaptively select the optimum disparity at each pixel among all scales.

## 5 ENCODER-DECODERS FOR DEPTH ESTIMATION

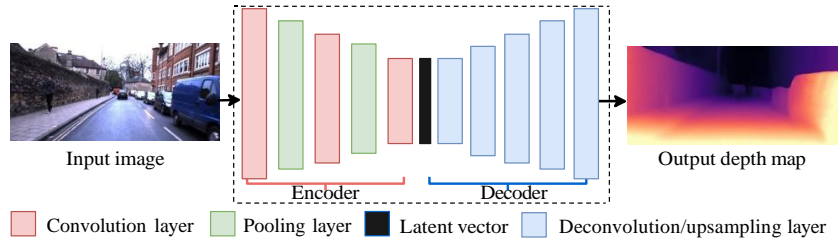

Fig. 4. Baseline CNN encoder-decoder network.

The existing deep learning models generally follow a common architecture with an encoder-decoder design. The encoder is also known as the backbone and acts as the feature extractor. The decoder produces the depth from the extracted features by upsampling the encoded feature vector from a low to high-resolution depth map. Therefore, we further decompose the non-linear processing function $f$ of Equation (1) into two stages: encoding and decoding. That is:

$$d = g(h(I)) = (g \circ h)(I). \tag{3}$$

In other words, $f$ is a composition of two functions $h$ and $g$. The function $h$, referred to as the encoder (Section 5.1), maps the input $I$ into a latent space, also called a latent vector, while $g$, referred to as the decoder, decodes the latent representation into a depth map (Section 5.2). Decoding usually results in a coarse depth map $d$, which is further processed with a sequence of blocks, hereinafter referred to as refinement blocks, to improve its accuracy and spatial and depth resolutions.

This section briefly discusses the general encoder-decoder design (Figure 4). The section also describes the different design choices that existing methods use on top of the general encoder-decoder to achieve better accuracy. We identify three such common design choices. They are (1) the use of multiple encoder-decoders, (2) multi-scale feature fusion (Figure 5) within encoder and decoder, and (3) multi-scale depth estimation (Figure 6) at the decoder.

### 5.1 Encoder

Existing network architectures implement feature extraction as an encoder $h$ or an encoder followed by a set of fully connected layers. A basic encoder has a set of convolutional layers to downsample the input image to a latent space. However, using a stack of convolutional layers only is a computationally costly process that results in an extensive number of parameters to learn. Adding pooling layers in between helps to downsample the outputs of convolutional layers and reduce the dimensions. Thus, they help to reduce the number of parameters. Therefore, a typical encoder consists of a stack of convolutional and pooling layers. The performance on the monocular depth estimation task also depends on the number of layers within the encoder because as the number of layers increases, the network gains higher representation ability [96]. In CNN-based architectures, the encoder can be fully convolutional (VGG and AlexNet) or a network with a set of residual layers with skip connections such as ResNet [43]. Such residual networks (e.g., ResNet18, ResNet30, and ResNet50) perform better for depth estimation tasks [74] than fully convolutional ones because they are easier to train, especially for deeper architectures (architectures with a higher number of layers). For instance, ResNet50 achieves the best results [43, 155]. ResNet architectures pre-trained on Imagenet perform better compared to the ones that are not pre-trained. However, residual nets with more parameters lead to overfitting due to catastrophic forgetting with longer training periods [45]. Therefore, the performance of pre-trained ResNet architectures also degrades with more parameters.

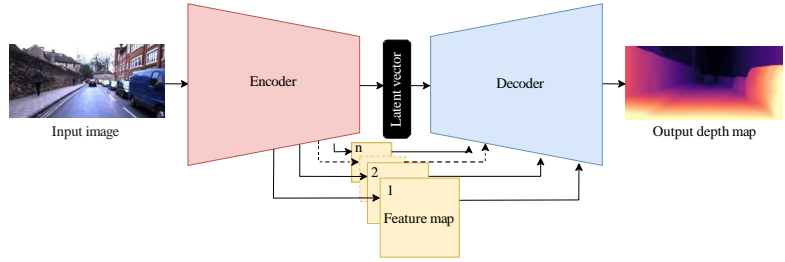

Fig. 5. Extended encoder-decoder with multi-scale feature fusion [57, 172, 176].

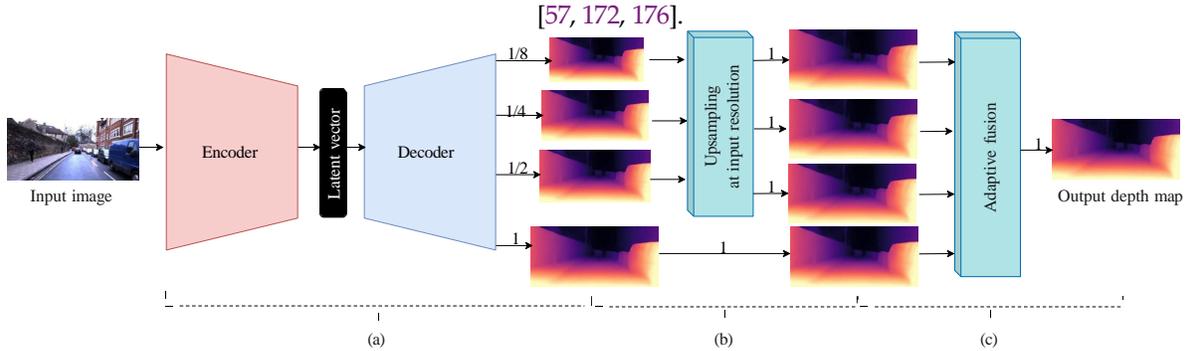

Fig. 6. A generalized pipeline for the three multi-scale depth estimation strategies at the decoder. (a) Estimating multiple scales of depth outputs using encoder-decoder [114]. (b) Upsampling the estimating depth at multi-scale to a single scale [44]. (c) Fusing depth estimations to the original scale with weighted summation [185].

Some methods use Vision Transformers (ViTs) as the backbone to implement attention mechanisms and enable the network to focus on specific areas of the image or video [8, 16, 28, 125]. Moreover, when dealing with video input, existing works often use RNN, such as LSTM networks, to ensure temporal consistency in the reconstruction. LSTM also enables the network to base its current predictions on what it has previously predicted. Therefore, LSTMs are more suitable for time-series inputs. We describe these methods in Section 6.2.

## 5.2 Decoder

The decoder module takes the latent output vector from the encoder (Figure 4), processes it, and produces the target output, depth. Therefore, the decoder is responsible for up-sampling the latent vector, and it includes mainly unpooling, up-convolutional, or fully connected layers for mapping the features to the predicted depth map output. Several methods in the literature adopt the decoder design of DispNet[104] with additional upconvolutions to make the results smoother. DispNet estimates depth in a single scale and later methods propose to modify their decoder architecture to use multi-scale depth estimations (see Section 5.5). Many models use ResNet-based decoders as well. An ablation study [114] shows that using ResNet50 as the decoder significantly improves the accuracy compared to ResNet18; the decoder architecture of MonoDepth2 [44].

## 5.3 Multi-task learning

Many methods consider depth as part of a multi-task learning problem [60, 136, 158]. Multi-task learning involves estimating multiple modalities such as depth, optical flow, surface normals, and object-level segmentation. For instance, Saxena et al. [136] use a single encoder-decoder network for optical flow and depth estimation.

Using multiple encoder-decoder networks in parallel, rather than a single encoder-decoder, is also a popular approach. Multiple encoder-decoders are particularly efficient for multi-task networks. Such multi-tasking methods train each encoder-decoder block separately or jointly. The latter is motivated by the fact that these tasks are interrelated and can help improve the performance of each other if trained jointly [60, 158]. For example, DeMonNet [154] proposes a chain of five encoder-decoder networks to estimate optical flow, ego-motion, coarse depth, and refined depth using an iterative training approach. Several methods [58, 98] exploit the shared encoders and decoders between tasks for multi-task learning. Poggi et al. [120] experiment with two decoder networks and a single shared encoder. The conclusion is that multiple decoders show better performance than a single decoder. However, they require more memory and processing at runtime.

## 5.4 Multi-level feature fusion

Most depth estimation methods focus on feature maps of multiple scales to achieve higher accuracy [7, 57, 92, 95, 177, 191, 197, 210]. The basic idea is to aggregate hidden features at lowers scales with the feature maps at a higher scale. Thereby, methods upsample the low-texture regions and blurry fine details to gain high accuracy. Therefore, each upsampling stage in the decoder consists of upsampling and concatenating layers to fuse low-level features from the encoder. Figure 5 shows the multi-scale feature fusion between the encoder and the decoder. Similarly, Chen et al. [24] adopt an hourglass architecture to obtain multiple-scale information. Here, skip connections concatenate the features. Multiple encoders also enable multi-scale feature fusion to obtain features of input images at different scales [176, 185, 205]. Xu et al. [176] propose a multi-scale feature fusion dense pyramid for this purpose. They use attention mechanisms to adaptively fuse global and local features of input images at multiple scales. Figure 5 shows the pipeline of feature fusion using single and multiple encoders.

## 5.5 Multi-scale depth prediction

Many existing methods compute depth at different scales at the decoder level. The multi-scale depth estimation strategy generates high-resolution depth maps from low-resolution outputs in a coarse-to-fine fashion. Eigen et al. [36] exploit this idea by estimating multiple depths at different scales. They compare the output depth to its ground truth at the corresponding scale. Finally, they calculate the overall loss by combining the individual losses at each scale within the decoder. Depth at multiple scales lowers the likelihood of the model converging to a local minima [44]. Therefore, it enhances the training efficiency.

Figure 6 shows a general pipeline for the three strategies to employ multi-scale depth estimation. One way is to produce depth maps as side outputs at each upsampling stage of the decoder and compute the loss separately at each scale. For example, self-distillation-based CNN architecture by Petrovai et al. [114] estimate four depth maps at four scales ($1/8\times$, $1/4\times$, $1/2\times$, $1\times$) as outputs. The second group of methods, first estimates depth at multiple scales. Then they upsample each scale to a single scale to compute the loss [44] only at a single scale. For example, the decoder estimates depth at three stages ($1/8\times$, $1/4\times$, $1/2\times$). The upsampling stage up-scales the depth maps separately to obtain four $1\times$ depth maps before computing the loss. The third group of methods uses depth fusion blocks to adaptively fuse the depth maps with different scales to obtain a single fine-tuned depth map and compute the loss [172, 185]. Such methods deploy CRFs or attention mechanisms for adaptive fusion.

## 6 NETWORK ARCHITECTURES

We categorize the methods based on the structure and composition of the networks. We focus on CNN, Random fields, Recurrent Networks, Generative Adversarial Networks, and Vision Transformers as the main types of architectures. Additionally, we describe hybrid architectures and their advantages. Section 6.1 explains the CNN-based encoder-decoder network architecture as the baseline architecture. The rest are architecture variants. Section 6.2 compares the architecture variants concerning their advantages and limitations over CNNs. We categorize the papers selected for the survey according to their architectural similarity.

### 6.1 CNNs as the baseline

The encoder in a CNN is responsible for downsampling the input image while aggregating features so that the decoder can later use the output of the encoder network to generate the depth maps. Figure 4 depicts a schematic encoder-decoder network. CNNs are very powerful in modeling local features. Increasing the number of hidden layers in the network leads to a large number of parameters. They estimate the desired depth map by learning the non-linear parameters of the local regions of the input [72, 168]. However, recent experiments reveal that using conventional CNN operations to extract features often results in blurry predictions [155]. This is because the vanilla CNN operations consider every pixel as a valid pixel for feature extraction. Although this enables extracting the majority of the valid features that represent the monocular cues that are present in the input image, they also encode the unnecessary features of invalid pixels. Gated convolutions, on the other hand, are capable of switching off the invalid pixels from further processing through the encoding process. Thus, enabling the dynamic feature selection for each channel across all encoder layers. Verie et al. [155] use a standard CNN coupled with gated convolutions in their architecture to produce sharper and better depth maps. The feature fusion of a conventional CNN decoder can be described as a function of $g$ [58].

$$g = \theta\left(conv\left([x, F]\right)\right). \tag{4}$$

Here, $\theta$ is an activation function and $conv$ is the convolution operation of the current layer. The decoder fuses the upsampled decoded feature $x$ from its previous level and the corresponding encoder feature $F$ through a concatenation layer followed by a convolutional layer (Figure 5).

Detail loss is a common issue in conventional CNN encoders [53]. Detail loss happens because of the intensive number of strided convolutions and spatial pooling. It causes the models to output low-resolution depth maps by reducing the resolution and spatial information during encoding. However, using dilated convolutions for encoding reduces this effect as they enable extracting more contextual features while retaining the spatial resolution of the feature maps produced by the encoder [50]. They also improve performance by reducing the number of computations and increasing the receptive field.

*6.1.1 Fully convolutional networks - FCN.* Architectures with fully connected layers used for the decoder are computationally expensive as the neurons in fully connected layers are connected to all neurons in the preceding layer. This results in a large number of parameters [82]. However, Fully convolutional networks (FCNs) help eliminate or reduce dense layers to minimize the number of learnable parameters in CNNs. Generally, they make use of $1 \times 1$ convolutions instead of fully connected layers, allowing the network to train on variable input sizes. Monocular depth estimation also benefits from FCN-based architectures, which are fast training models with a variety of up-convolution and skip connection layers [20, 43, 50, 78, 189].

*6.1.2 U-Net.* Many depth estimation methods use U-net as a common FCN-based encoder-decoder architecture. The general U-net architecture consists of multiple encoders and decoders with skip connections, which are connected to create a U-shaped architecture. Most existing monocular depth estimation networks use the U-net architecture [44, 65, 155, 165] and its variations. The number of layers used in the U-net architecture depends on specific requirements such as intended training and inference resolution [106]. A key advantage of using U-net over other designs is its ability to represent abstract features (contextual information) as well as local information. However, when the U-net becomes deeper (i.e., with more layers), the network requires long connections to bridge the corresponding encoding and decoding layers. In such cases, having long connections causes the U-nets to produce an output with information gaps, such as missing fine details, as the recovery of the encoded information at the decoder is difficult [82].

*6.1.3 3D-CNN.* The general CNN network architectures discussed above execute the kernel operations in two dimensions only (RGB images with three channels). Some studies have also focused on using 3D convolution-based network architectures (3D-CNN) to extend the boundaries of using 2D convolutions. As opposed to 2D convolutions, 3D convolutions slide the kernel in three dimensions. A sequence of striding and pooling operations in 2D convolutions decreases the model's performance for tasks requiring fine-grained details. Therefore, traditional encoder architectures fail to propagate enough details for the decoder, so conventional upsampling operations at the decoder can recover accurate depth. To overcome this limitation, Guizilini et al. [45] propose to replace 2D convolutions with 3D packing and use unpacking instead of upsampling in the decoder. They present a near-lossless encoder-decoder using the packing and unpacking operations. However, packing and unpacking operations increases the network complexity as they result in more parameters. For example, the model by Guizilini et al. [45] reached around 128$M$ parameters.

Apart from 2D images, some methods also use temporal data as an additional dimension. Zhang et al. [195] show that 3D-CNN helps capture the change and motion information in consecutive frames for monocular depth estimation. They improve the architecture's performance by convoluting the input along both the spatial and temporal dimensions. Temporal relations are most suitable for real-time depth estimation as they continuously require RGB video sequences during execution.

## 6.2 Network architecture variants

In this section, we discuss other architectures as variants to the baseline CNN and compare the advantages and disadvantages of them.

*6.2.1 Random Field-based.* As mentioned in Section 6.1, CNNs perform well in extracting local features rather than in global (contextual) features. Compared to the baseline CNN model, random fields focus more on these global features. In monocular depth estimation, the features around the local neighborhood are essential. However, the relations between the neighboring pixels and the relations between the neighboring regions (patches) or superpixels also carry essential information. Such global features provide information related to the depth of a particular region of an image. Probabilistic models (e.g., random fields) are good at modeling global features [135].

Random fields are probabilistic graphical models that are competent in taking the relations of a particular pixel to its neighboring pixels or superpixels. These methods create graphical models to represent complex structures using pixel values as nodes and their neighboring pixel values as edges. Early attempts in depth estimation leverage this nature of random fields [135]. Therefore, random field models consider statistical properties that encode prior assumptions and enable parametric learning. Consequently, the estimated depth is statistically similar to the distribution sample [70]. These graphical models can learn only limited features compared to deep learning models, which can learn a large number of parameters to extract features. Two types of random fields extensively used are Markov Random Fields (MRFs) and Conditional Random Fields (CRFs).

**(i) MRF.** MRFs enable the use of more contextual features and reasons more globally about the spatial structure of the scene. Saxena et al. [134, 135] consider the depth smoothness of nearby pixels or superpixels (patches). Later work on hierarchical multi-scale MRFs shows that MRFs can also model the depth of a patch of an image, its neighboring patches, and other possible patches that are not close to each other. Followed by this work, many methods use MRFs for depth-related tasks [73, 108].

**(ii) CRF.** CRF is a special case of MRFs that produce a structured output. CRFs model the interactions between pixels or superpixels for depth estimation. A significant amount of research formulates depth estimation as a conditional learning problem [94, 211]. The three main types of prevalent CRFs in depth estimation are continuous, discrete, and hierarchical CRFs. Liu et al. [90] propose a continuous CRF, while Liu et al. [94] consider both continuous and discrete depth information. The latter uses continuous variables to represent depth within superpixels and discrete variables to represent interactions between superpixels. Zhuo et al. [211] adopt a hierarchical CRF method and formulate the depth estimation not only using superpixels but also with regions and the overall layouts of the input image. Hierarchical approaches perform well since they also provide the benefit of modeling global context using pairwise relations. However, CRFs are slower to train compared to vanilla CNNs.

*6.2.2 Recurrent Network-based.* Recent monocular depth estimation methods often make use of the temporal behavior of the sequential data (e.g., methods that rely on monocular videos for supervision). Although convolutional-based auto-encoders are capable of mapping the local relations of the features for depth estimation, they lack the ability to model the time series information. Any information that was present in the past data of the sequence is hardly taken into account by CNN architectures. However, the sequential data of the monocular videos possess useful information for the depth estimation task, from which the networks can learn the dependencies of the depth changes of objects in the scene across time. Recurrent-based architectures play a prominent role in this scenario. Recurrent networks consist of cyclic connections that enable the neural networks to learn the temporal behavior of the sequence of frames of a video [162]. Hidden states of a recurrent network act as agents in storing the sequence relations in their internal memory for future use. Compared to CNNs, an RNN back-propagates the information flow over time across previous instances.

**(i) Traditional RNNs.** Traditional RNNs contain only hidden states, which store the memory of the previous state. Therefore, when processing, they take both the current and previous states. Vanilla RNNs can learn while back-propagating the gradient in time. Inspired by traditional RNNs, Gehgrig et al. [41] propose a generalized RNN, which handles data from multiple sensors. However, vanilla RNNs lack persistence in long-range sequences as they cannot store the previous state for very long.

**(ii) LSTM-based.** LSTM networks learn long-term dependencies of multiple instances using both hidden states and cell states. Zhou et al. [209] use an LSTM decoder, which ensures that the memory has a higher impact on improving the predicted depth accuracy. Compared to fully connected LSTMs, convolutional LSTMs (convLSTM) are better for depth estimation since they not only enforce the learning of appearance cues at multiple scales but also take temporal data into account [102, 118, 195]. Kumar et al. [31] and Wang et al. [162] show that a recurrent-based architecture with LSTM units is capable of achieving better results with temporal information obtained from multiple consecutive views of a video. They both use convLSTM units to incorporate the motion and temporal data in previous frames to estimate depth in the current frame. However, as per the results of the ablation study from Wang et al. [162], interleaving LSTM units across the encoder performs better than interleaving across a full network or only the decoder (Figure 7).

Similar to CNNs, in some cases, recurrent networks also leverage attention mechanisms to enhance performance by leveraging long-range dependencies.

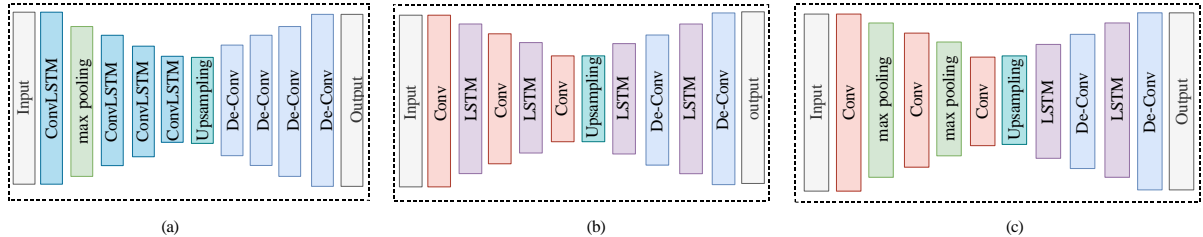

Fig. 7. LSTM layer arrangements in baseline architecture. (a) Convolutional LSTM layers in encoder only. (b) LSTM layers in both encoder and decoder. (c) LSTM layers in decoder only.

*6.2.3 GAN-based.* A Generative Adversarial Neural (GAN) network consists of two networks: a generator and a discriminator. The generator aims to produce realistic data from a random vector, and at the same time, the discriminator classifies the data as real or fake. In this context, a generator is responsible for producing one image (e.g., depth image, RGB image) from a given input vector, and the discriminator is responsible for determining the resemblance of the generated image with its original image (e.g., original depth image, original RGB image). Recent depth estimation methods use GAN networks due to their ability to generate depth images with higher accuracy. Unlike the baseline CNN model, GANs can generate a plausible distribution of samples to render an image. GANs are beneficial in modeling a more realistic depth image or another view. Therefore, monocular depth estimation techniques deploy the generative behavior of GAN networks for image or depth reconstruction. There are two modes of GAN in this context,

· **GAN for depth map reconstruction**: In this mode, given the RGB image and its corresponding ground-truth depth during training, the generator network learns to predict a depth map similar to the ground-truth and the discriminator verifies the similarity of the generated map with the original ground-truth. The discriminator is responsible for distinguishing the synthetic from the original depth maps.

· **GAN for image reconstruction**: In this mode, the network learns depth as an intermediary variable using multiple views of the same scene. The generator attempts to reconstruct one view from another as input and the discriminator verifies the generated view with the original target view [2].

**(i) Vanilla GAN.** Vanilla GANs cover the basic adversarial network architecture with a generator and a discriminator. A well-trained discriminator to distinguish the dissimilarity between the rendered and original image is likely to produce better results. Several studies use vanilla GANs with adversarial training for monocular depth estimation [2, 32]. In Kumar et al. [32], two subnetworks, DepthNet and PoseNet, jointly form the generator network. They predict the depth and pose estimation separately in rendering a pair of predicted images so that the discriminator can score the likelihood of the similarity between the original and rendered images. They use skip connections in a general encoder-decoder-based architecture design to reduce the problems caused by vanishing gradient problems. Zhao et al. [203] and Xu et al. [178] also propose similar approaches with multiple subnetworks for depth, pose, and ego-motion. However, one major issue in vanilla GAN is when the generators produce over-optimized plausible depth outputs that agree with the discriminator. Therefore, the generators produce a smaller variety of outputs. This limitation is called the mode collapse [204].

**(ii) Cycle GAN.** Several studies use Cycle GANs in monocular depth estimation due to their ability to achieve higher performance in image translational tasks. The major advantage over vanilla GANs is that Cycle GANs allow the geometric consistency between the input and target images to generate high-quality depth maps. For a GAN to become a Cycle GAN, it must have at least two GAN networks as subnetworks. Zhao et al. [204] propose a symmetrical network architecture to generalize real and synthetic domains for depth estimation. Their network learns to generate synthetic-to-real and real-to-synthetic images, preserving geometric and semantic content. On the other hand, Pilzer et al. [117] propose to deploy two GANs to reconstruct two distinct disparity images from stereo image pairs that mutually constrained each other in a cycled manner. Although cycled GANs successfully translate inputs, the computational time taken for training is higher than vanilla GANs.

**(iii) Conditional GAN.** Conditional GANs (cGANs) are GANs that condition some additional information. They enable the conditional generation of images using auxiliary information. The purpose of using auxiliary information as conditions for monocular depth estimation is to stabilize the vanilla GANs by diminishing issues (e.g., gradient artifacts). Chen et al. [22] propose a constrained GAN by applying spectral normalization for the generator and discriminator. They use an adversarial-trained U-net that uses depth cues and context-aware structural details to estimate depth. Similarly, Jung et al. [66] propose a batch normalization-based cGAN.

GAN-based networks generate depth maps while preserving structural information. However, the general performance of the GAN-based architectures highly depends on the accuracy of the discriminator. Any adversarial examples with missing regions (due to corners in the image and motions) can cause the discriminator to predict incorrect results, and the final rendered depth images may not look realistic due to this limitation. Trying to eliminate such cases can slow down the overall learning rate of the network [32].

*6.2.4 ViT-based.* While convolution-based architectures yield good results in monocular depth estimation, they often exhibit limitations in modeling long-range dependencies due to the locality of the convolutional operation [182]. Handling long-range dependencies with LSTMs is also challenging as the LSTMs only take small segments (few consecutive frames) of video input independently. Several methods [8, 16, 28, 125, 182] propose ViTs to address such limitations. ViTs represent an image as a series of patches. They are known to be a better solution than the CNNs as they overcome the limited receptive field issue.

In an encoder-decoder network, the backbone encoder for the model, has a higher influence on the capabilities of the model. For instance, when the encoder is a CNN, the down-sampling operations may lose important information from the original input. Although down sampling progressively increases the effective receptive field when it goes deeper in the network, it will be hard to regain the lost features from the decoder. ViTs outperform CNNs because of their nature of embedding images while maintaining a constant dimensionality throughout all the processing stages of the encoding [182]. Additionally, compared to CNN and recurrent networks, ViTs assume minimal prior knowledge [68, 110].

ViTs have achieved better results in depth prediction with fine-grain details using the global receptive field. ViT-based models are shape-biased, while CNN models can be more texture-biased. In their experiments, Bae et al. [8] show that shape-biased ViTs better generalize over depth than texture bias (e.g., CNN) ones.

The total learnable parameters are high with ViTs. However, due to its parallel processing capabilities, the training process does not severely impact the computational time during training. ViTs enforce the information embedding globally across the overall image using self-attention blocks. However, one drawback is the computational complexity. The complexity of the models depends on the performance of the self-attention blocks. However, the input image resolution has a higher impact on the performance of the self-attention blocks and predicting pixel-wise depth at a higher resolution is challenging when ViTs are used [28].

*6.2.5 Hybrid architectures.* In this section, we discuss several hybrid methods that leverage the strengths of different architectures together to improve the overall performance of their framework. CRFs are most commonly used as a refinement step with other networks such as CNN and GANs. A recent trend is to use ViTs with CNN as hybrid architectures.

**(i) CNN-CRF.** Since CNNs utilize local features and CRFs, on the other hand, are proven to be best with global (contextual) features, existing hybrid methods use them together for monocular depth estimation. Most methods use CRFs for the predicted depth as a post-processing step to refine the output. The output depth maps are sharper and more edge-conforming in hybrid CNN-CRF-based methods [19, 83, 91, 144, 172–174]. In Song et al. [144], propose a contextualized CNN by using class level and pixel level refinement through CRFs. Xu et al. [172] propose a multi-scale fusion-based depth estimation guided by CRFs.

**(iii) GAN-CRF.** GANs are suitable for handling under-fitting-based issues in models [32]. CRFs, on the other hand, can be utilized for the structural refinement of depth maps. Puscas et al. [122] propose CRFs to couple the outputs of the generator network and the discriminator network of a GAN to ensure the structural consistency of the output in an end-to-end way.

**(iv) ViT-CNN.** Using ViT alone is challenging for predicting continuous dense outputs like depth. Since transformers provide a self-attention mechanism, several methods propose hybrid architectures that use both CNN and ViTs [6, 182, 197]. Yang et al. [182] propose a hybrid network with ResNet50 followed by 12 transformer layers as the backbone for the encoder and extract features of 1/16th of input resolution. The results are still better than those of using CNN encoders alone. With a similar architecture, Bae et al. [8] enable higher accuracy for fine details by using a hybrid ViT-CNN architecture.

## 6.3 Attention mechanisms in architectures

Humans do not process everything in their field of view at once. Instead, they possess the ability to focus only on a subset of information. This process of selectively concentrating on the most important information enables perceiving things faster and more accurately. Deep learning methods incorporating attention mechanisms during training [111] utilize this idea for computer vision tasks. Conventional deep learning models treat all the features extracted from the input similarly. However, some information may have a higher impact on the classification or regression task, while others may not. Attention mechanisms focus on more important features for processing while suppressing unnecessary ones [167].

In this context, there are features that influence depth estimation more than others. For instance, the depth of a pixel does not depend only on the nearby pixels. There can be pixels widely spread in the input, but they carry useful information for the depth of a particular pixel. An attention mechanism is usually adopted to extract such long-range global dependencies. Moreover, general loss functions used for 3D supervised or self-supervised learning methods assume that all the regions in the input image contribute equally to the depth of the scene [64]. In contrast, an attention-based loss function focuses on what is necessary. Based on that, Jiao et al. [64] propose a novel loss function, which leverages depth awareness using attention. Following their work, SharpNet [124] also proposes an attention-based loss function that focuses on the occluded contours while training.

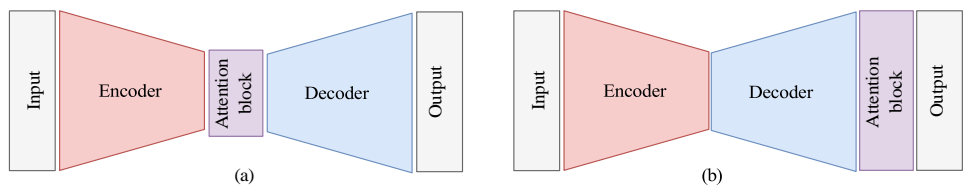

Fig. 8. Attention blocks in encoder-decoder architecture. (a) Attention before decoding. (b) Attention after decoding.

Attention maps generated during the training process guide reasoning on which information is more relevant for the final depth output. Several methods [89] use attention mechanisms to fuse depth at multiple scales. Using attention for training boosts depth estimation with respect to meaningful semantic information [147].

The improved version of the baseline architecture contains three components: encoder, decoder, and attention block (Figure 8). The order of arranging these components can differ. One way of arrangement is to encode the input image, compute the attention maps and then decode the latent vector [65]. The other way is to encode first, then decode, and finally apply the attention to the higher resolution tensors [13]. Besides, ViTs (Section 6.2.4) also employ attention mechanisms. There are three ways to employ attention: spatial, channel-wise, and mixed.

*6.3.1 Spatial attention.* Spatial attention is about where to focus, allowing the networks to learn the locations with important features. It transforms the spatial information in the input image to another space while retaining key information [111]. In other words, it encourages attention to more important spatial positions while training. Several methods use spatial attention for monocular depth estimation [64, 65, 174, 177, 196, 206].

Jiao et al. [64] show that the depth estimation training process encounters a data imbalance issue due to a long tail distribution of depth values in indoor and outdoor datasets. Because of this, easy samples with small depth values contribute more to the output than hard samples with large depths. Therefore, models overfit to predict small values. They propose to pay attention to larger depth values using a spatial attention-driven loss function. Johnston et al. [65] propose a self-attention module to estimate the similar depth in non-contagious regions (global). Since many of the existing supervised and self-supervised methods do not consider global context, they argue that using self-attention helps to handle long-range dependencies in monocular depth estimation.

Xu et al. [174] use a spatial attention mechanism at multiple scales to control the amount of information processed in corresponding features at different scales. As a result, they utilize a structured attention module to identify the information pertinent to the final depth automatically. Similarly, Xu et al. [177] propose a multi-scale spatial attention-guided approach with semantic enhancement.

*6.3.2 Channel-wise attention.* Channel-wise attention networks learn what information to focus on [111] regardless of the spatial location. Having multiple channels (multiple filters) CNNs allow the network to extract more information by learning distinct features with less computation time. However, not all channels equally contribute to the depth of a scene. Architectures use channel-wise attention mechanisms to pay attention to the more important channels during training [82, 84, 145, 174, 179].

Accurate overall scene structure and relative depth are essential for many applications, such as autonomous driving and augmented reality. Most existing methods lack explicit modeling of the global scene structure. Channel-wise attention allows the neural networks to aggregate global features. Aggregation enables the network to emphasize informative features that contribute to the overall scene structure rather than its local features, such as neighboring pixel values. Yan et al. [179] show that an ordinary decoder does not preserve the fine details when it concatenates high-level features with low-level features. This results in blurry artifacts in the depth output. Therefore, they propose a channel-wise attention-based network that provides an overall structure perception and emphasis on details. To this end, they employ self-attention to capture long-range dependencies and a weighted summation of the aggregated features. The methods show better results with sharper object boundaries in the final depth output.

*6.3.3 Hybrid.* Some methods adopt a combination of spatial of channel-wise attention mechanisms [28, 59, 81, 109, 190]. For instance, Naderi et al. [109] argue that conventional encoder-decoder networks do not consider the similarity between the RGB image and its corresponding ground-truth depth. Therefore, they propose to treat the similarities as a constraint in estimating the depth. Thus, they enhance the generic encoder-decoder architecture with an attention module that incorporates spatial and channel-wise attention. Their lightweight, adaptive geometric module can measure such similarities using cross-correlation between the encoder and decoder.

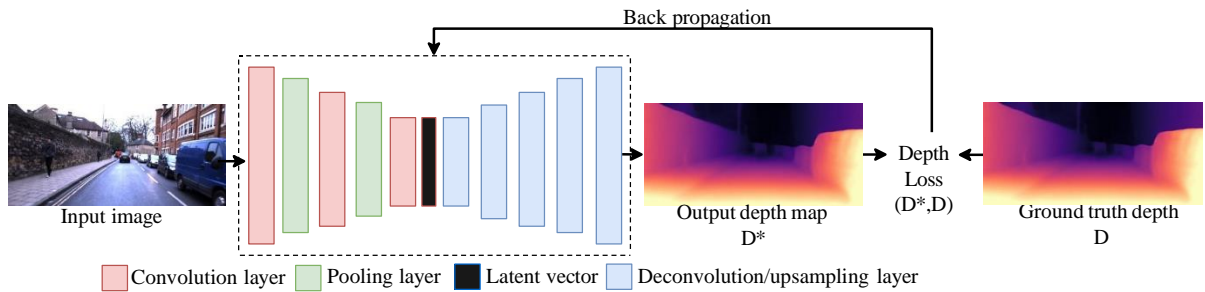

Fig. 9. General 3D supervised training procedure.

# 7 DEGREE OF SUPERVISION

## 7.1 3D-Supervised methods

Supervised methods regress the depth directly from monocular images by obtaining the supervisory signal from ground-truth depth, ground-truth disparity, or manually annotated depth labels provided at the training.

Most methods use the depth acquired by range sensors as the supervisory signal during training. Such trained models are then directly used to predict the depth of each pixel during testing. The models perform well under different datasets, training to learn depth related to objects and shape priors. Models aim to minimize the loss between the predicted and the ground-truth depths. Figure 9 shows the generic 3D supervision procedure.

*7.1.1 3D Supervised Methods: From Early Methods to Current Trends.* We divide the main 3D-supervised methods into two categories based on whether they rely on hand-crafted features.

Saxena et al. [134] proposed the first supervised monocular depth estimation method that considers depth at individual points and the relation between depths at different points using MRFs. Later work on enhancing this method is with a patch-based supervised model for unstructured environments [135] that relies on hand-crafted features [135]. The improved method considers monocular cues from local patches and global optimization of the final results for the depth relations using MRFs.

Karsch et al. [67] propose a depth transfer-based framework that relies on non-parametric depth sampling. However, during runtime, retrieving depth information from a large collection or a database of RGBD images to search for candidate images resembling the input image is computationally expensive. They assume that similar images are more likely to have similar depths. Therefore, early supervised methods extensively employed hand-crafted features to capture the correlation between a color image and its corresponding ground-truth depth.

Deep learning-based methods produce accurate results (Section 6) due to their ability to train end-to-end for multiple tasks (optical flow, surface normal, depth). Eigen et al. [36] use CNNs to regress depth maps. In line with the current trend, numerous studies focus on enhancing network-based methods [35, 80, 132, 201]. These methods obtain depth maps directly as the output from the decoder network or as an average estimate over a depth distribution [168]. Section 7.5.1 further describes the 3D-supervised-based loss functions.

The major limitation is that these methods rely on the availability of large collections of ground-truth depth data. They have a limited capability in generalizing to previously unseen data [154].

## 7.2 Self-supervised methods

Self-supervised monocular depth estimation methods tackle the challenge of not having enough ground-truth 3D data for training. These methods achieve promising results by formulating the problem as an image reprojection problem. During training, these methods utilize depth as an intermediary variable and generate a targeted view by reprojecting a source view using image synthesis techniques and the predicted depth. These deep learning models learn to estimate depth by minimising the dissimilarity (similarity loss) between the synthesized (reprojected) view and the target view. Image reprojection and loss minimization are the two main steps involved during training. These methods require more than one view of the same scene as the input during training, and one of those views is taken as the target image to obtain the supervisory signal. Figure 10 (Section B in Appendix A) illustrates the generic self-supervised training process.

We further categorize the self-supervision methods as stereo-image-based self-supervision and monocular-video-based self-supervision.

*7.2.1 Monocular depth by stereo supervision.* These methods use rectified stereo image pairs as training data to learn depth by estimating the pixel disparities between the image pairs. Still, they are capable of estimating depth from monocular images during testing. In other words, during training, these methods either use the left image to obtain an inverse warp of the left image by applying the predicted disparity on the right image [39] or use predicted left disparity for the input left image to warp the right image to obtain the left image [11, 43, 189, 194]. For example, Garg et al. [39] estimate a single disparity per pixel by computing the photometric difference between the source-left image and the inverse warped target image. The method produces the target image by applying the predicted depth on the right image. For accurate disparity maps, Godard et al. [43] train the network to predict disparities aligned to both the left and the right images. However, they provide only the left view as input to the convolutional layers of the network. They propose the L1 disparity consistency penalty to ensure coherence. This cost makes the left-view disparity map consistent with the projected right-view disparity map. During training, some methods use a discrete probability distribution of disparity for each pixel and the final value is obtained by taking the argmin [171].

These methods also have some common limitations. Some pixels or regions may present only in one image of the stereo pair due to occlusions existing in the scene (stereo-disocclusions). During image reconstruction, these pixels produce noisy images. Therefore, the methods are unable to produce disparity for such pixels. In such cases, some methods [44, 75, 208] use different masking strategies to ignore pixels that violate the assumption of no occlusions. Another limitation is that warping methods are prone to producing holes in the synthesized image, which causes inaccurate results. One way to mitigate this limitation is by using post-processing steps. Recent methods use multi-view stereo images to ensure robustness to noise during training [44, 75, 208].

Consistency through stereo image pairs can be challenging due to noise caused by illumination variations and low texture regions. Some methods mitigate such issues by using a stream of stereo imagery. However, in rectified stereo-image sequences, the effect of stereo-disocclusion (the regions visible in one image but not in the opposite image) is considerably high.

*7.2.2 Monocular depth by monocular video supervision.* While the previous category required stereo data during training, depth by monocular supervision requires only monocular image sequences and videos during training [208]. The distinguishing feature is their focus on the temporal or geometric consistency between the adjacent frames in a sequence of images. These methods assume that the scene is static without moving objects, with no occlusions between the target and source views. Unlike the previous methods, they estimate depth using either known camera parameters or by estimating camera pose and ego-motion through a separate network [44]. They also formulate the problem as minimizing appearance matching loss and camera pose error. They require one or more source images of a monocular video (different viewpoints) to train the view synthesis network. The network learns depth while inferring the explicit scene geometry and camera pose [208].

Since the input is multiple adjacent frames, the model takes the current frame as the target frame and the previous and next adjacent frames as the source images. While most methods use separate networks to estimate depth and camera pose, recent methods train these networks jointly. DepthNet (the depth network) takes the target frame and estimates the depth value and PoseNet (the pose network) takes the entire image sequence and predicts the camera pose from target to source. These methods reconstruct the inverse warped image by sampling the pixels from source images. Consequently, the adjacent frames provide the supervision signal for training.

During training, these methods also require different masking strategies for robustness towards occlusions [53], non-rigid motions [159], or stationary pixels [213] (for objects moving at the same velocity as the camera or when the camera is not moving). For instance, one of the main challenges in these methods is the presence of moving objects, which leads to inaccurate camera pose estimation. Also, these methods highly rely on the accuracy of the camera pose estimators, although they only require the camera pose during training to constrain the depth network. These methods assume the surface is Lambertian (surfaces with constant brightness) so that the photometric reprojection error is meaningful [208]. Additionally, training these methods in the presence of occlusions, uniform texture, or repetitive patterns is challenging, as it becomes difficult to propagate the correct signal [114] when the scene violates these conditions. For instance, moving objects or a steady camera can cause the models to fail to produce accurate results.

Several methods focus on reducing the flickering depth and inconsistent geometry issues due to dynamic objects, repetitive patterns, and low texture areas in monocular videos [15, 58, 99, 198]. Bian et al. [15] propose to maintain the depth consistency of two consecutive frames. The method estimates depth for one frame and reprojects the estimated depth using estimated ego-motion for the next frame. However, this method only masks out the dynamic regions. Therefore, their method does not explicitly address the problem of estimating depth for dynamic regions that cause flickering effects. Luo et al. [99] propose fine-tuning the depth network weights at test time with geometric constraints. The method extracts geometric constraints using optical flow and corresponding pixels at distant frames. Geometric constraints reduce the inconsistent regions in the output during test-time training. However, the method is only robust for short-term dynamic scenes. Zhang et al. [198] extend the previous work by explicitly modeling scene flow for dynamic scenes. This method maintains the consistency geometrically and temporally. Hui et al. [58] also mitigate the effects of object motion by estimating a 3D motion field of moving objects along with depth and ego-motion. However, these models are not robust to inaccuracies that occur in reprojection due to illumination changes and occlusions.

Section 7.5.2 further describes the reprojection-based loss functions for self-supervised training.

## 7.3 Semi or weakly supervised methods

Semi or weakly-supervised methods mitigate the limitations of self-supervised methods.

*7.3.1 Semi-supervised methods for depth estimation.* Semi-supervised learning-based methods retrieve both supervised and cues from labeled and unsupervised ones from unlabeled images. Several semi-supervised methods improve image reconstruction-based depth estimation using ground-truth depth as an additional supervision signal. They estimate the inverse depth using stereo image pairs and maintain the photo consistency of the synthesized and target views. Additionally, the model also compares the estimated depth with the ground-truth depth. Therefore, semi-supervised methods use hybrid loss functions of image reprojection loss and per-pixel depth loss during the training [3, 46, 76, 184]. In addition, Amiri et al. [3] propose to use the left-right consistency. Baek et al. [10] leverage the same concept. However, they use separate networks for supervised and self-supervised loss and train them in a mutually distilled manner. Benefiting from the adversarial learning, Ji et al. [63] also train a depth regression network with two discriminators. Their work only requires a smaller amount of image-depth pairs with sufficient unlabelled images for training. , The generator network generates a depth image for an input RGB image, which tries to deceive the discriminator pairs.

Aside from the mentioned methods, other semi-supervised approaches, which employ the general idea of training from labeled and unlabeled data for supervision, are also available in the literature. For example, Tian et al. [152] propose predicting depth from unlabeled images using a depth network while verifying the depth network's confidence against ground truth using a confidence map. Meanwhile, Zama et al. [193] employ a semi-supervised method using image reprojection loss aided by semantic information learned via pixel-level ground-truth semantic labels during training.

Semi-supervised methods are valuable when available ground-truth depth data is limited. Even a smaller amount of ground-truth depth data provides a sound supervisory signal for training. Cho et al. [29] propose a semi-supervised learning-based student-teacher learning. The method only requires a small amount of ground-truth depth data for training the teacher stereo network.

*7.3.2 Weakly supervised methods for depth estimation.* Weakly supervised learning methods are helpful when ground-truth depth data are available yet noisy and sparse [87]. Additionally, limited ground-truth can enhance the performance of self-supervised learning by using depth or other existing sparse annotated (labeled) data as hints. These sparse data serve as a weak signal during training. The photometric reprojection losses fail to find the optimum depth using a global minimum due to the absence of ground-truth depth. Benefiting from depth hints, Watson et al. [164] propose to enhance self-supervised work to escape the problem of minimizing towards a local minimum. They use a depth map generated by another stereo algorithm to serve as depth hints. Similarly, Tosi et al. [153] use proxy disparity maps generated by a traditional stereo-matching algorithm to perform proxy labels to guide self-supervised learning. Methods which utilize joint 3D, self, and weakly supervision techniques are also available. For instance, Ren et al. [128] utilize semantic features as weak labels to ensure the consistency of the estimated depth. Sun et al. [149] use a pre-trained depth network to obtain pseudo-depth labels and improve the self-supervised training process for dynamic scenes.

Another way of obtaining weak supervisory signals for learning is through annotation with relative depth information. While ground-truth depth data collection with metric data is cumbersome, annotating pair-wise relationships is much easier and provides relative depth values. Relative depth cues are also known as relative ordinal depth. Following this idea, Zoran et al. [212] train a network to estimate the ordinal relations of the point pairs and later aggregate the results to obtain the output of the entire scene. Considering the ordinal relative positions of spatially different points, Chen et al. [24] show that deep learning models can directly predict depth by training with relative depth annotations.

Although high-resolution RGB images are more accessible, high-resolution ground-truth depth maps are not as they are in lower resolution. However, low-resolution and noisy depth maps still provide additional information for supervision. Therefore, the weakly supervised method by Xu et al. [175] utilizes low-resolution depth maps to train the network using resolution mismatched data.

## 7.4 Online learning

Offline learning/batch learning only allows the network to learn parameters on the training data. The model does not optimize the parameters during inference; instead, it freezes them, which makes the models specific to the training domain. On the other hand, online learning or learning on the fly enables the models to learn even after their deployment for inference. Online learning is beneficial for real-time applications of depth estimation as it allows for adjusting quickly to new incoming data [195]. Therefore, several studies which exploit the advantages of using online learning are available in literature [20, 58, 75, 97, 103, 112, 157, 187, 195, 200].

Two challenges in modeling online learning-based monocular depth estimation methods: scale ambiguity in monocular videos and lack of geometric information [200]. The latter causes the models to overfit the most recent data due to rapid and continuous scene changes and forgetting the past. The online learning process involves two steps.

*7.4.1 Pre-training (offline learning).* In pre-training, the model first trains offline using the available training dataset. Therefore, the network optimizes the parameters based on the offline datasets. For example, Maslov et al. [103] use a 3D-supervised loss, $L_s$ function to train the model offline while Wagstaff et al. [157] used a photometric reconstruction loss $L_u$, for unsupervised learning. Zhang et al. [200] use both loss functions to pre-train the model before deploying it for online learning.

*7.4.2 Online adaptation.* As the next step, the model trains again during inference by updating the pre-trained parameters to adapt to online data. Adaptation gradually improves the accuracy of predicting depth for the current input. Some methods select a single image for online learning, while others select batches of images. However, compared to mini-batches, large batches take longer time to train. Therefore, selecting the most appropriate batch size is crucial. Towards this objective, Zhang et al. [200] propose a meta-learning (learning to learn) based method enforcing online learning with few samples. They use the following gradient descent, $\nabla$, for updating model parameters where $\theta^d$ and $\theta^p$ are parameters for depth net and pose net, $I_t$, is the current frame, and $I_{t-1}$ is the previous frame [200].

$$[\theta^d, \theta^p]_{t+1\ t+1} \longleftarrow [\theta^d, \theta^p]_t - \alpha \nabla_{\theta^p,\ d} L_u([\theta^d, \theta^p]_t, I_t, I_{t-1}). \tag{5}$$

However, the model will require a longer time to train if the models wait a long time to obtain online data. Therefore, existing online learning-based methods maintain a buffer to store online data as a good practice, [75, 187]. Another limitation is that only the models that do not allow catastrophic forgetting of previous knowledge will not overfit the most recent data [97].

## 7.5 Loss functions

*7.5.1 3D supervision-based loss functions.* Per-pixel depth loss is the basic loss function, which tries to minimize the error between the predicted and the ground-truth. The per-pixel depth loss functions differ based on how the distance is measured between the actual and the estimated depth. Given the ground-truth depth $d_i$ and the estimated depth $d_i^*$ for pixel $i$ of $n$ number of pixels in the image, the per-pixel loss in terms of absolute loss $l_1$ can be formulated as

$$l_1(d^*, d) = \frac{1}{n} \sum_i^n |d_i^* - d_i|. \tag{6}$$

In $l_1$ loss, the sum of all the absolute differences between the ground-truth and the predicted depth values of pixels is minimized. Some methods have adopted $l_1$ based loss function [50, 154]. $l_2$ loss function, on the other hand, takes the summation of the squared differences [83, 134, 172].

$$l_2(d^*, d) = \frac{1}{n} \sum_i^n (d_i^* - d)_i.^2 \tag{7}$$

The $l_2$ metric has the risk of getting larger errors in the presence of incorrect predictions or outliers as it takes the squared differences of the values. Huber loss takes the best of both worlds. Huber loss takes the squared loss ($l_2$) if the error value is in a defined range or use $l_1$ otherwise. Therefore, Liana et al. [78] and Pintore et al. [118] use the reverse Huber Loss (BerHu) in this context.

Another loss function is the Scale Invariant Loss (SIL), which helps measure the relationships among data points in the scene. Measuring the relations among pixels irrespective of the absolute global scale helps to mitigate the depth ambiguities occurring due to the global scale of the estimated depth map output of the scene [36, 115]. SIL for a pair of pixels $i$ and $j$, $d$ as the log value of the difference between the per-pixel prediction and the ground-truth [36] is,

$$SIL = \frac{1}{n} \sum_i d_i^2 - \frac{1}{n^2} \sum_{i,j} d_i d_j. \tag{8}$$

Although the SIL relates to the $l_2$ metric, it is responsible for ensuring that the differences for the pixels are in the same direction and consistent with each other. Therefore, $SIL$ reduces the effect due to outliers, so imperfect predictions (outliers) will now output lower error values than $l_1$.

Many existing 3D-supervised methods use an enhanced scale invariant loss function introduced by Eigen et al. [35]. The second term in the new SIL is responsible for the first-order matching of image gradients of the predicted and the ground-truth depth maps. It ensures that the predictions are not only changing in small ranges but are also consistent in local structure. The improved SIL is,

$$ImprovedSIL = \frac{1}{n}\sum_i d^2 - \frac{1}{2n}(\sum_i d_i)^2 + \frac{1}{n}\sum_i [(\nabla_x d_i)^2 + (\nabla_y d_i)^2], \tag{9}$$

where $n$ is the number of valid pixels, $\nabla_x d_i$ and $\nabla_y d_i$ as the horizontal and vertical image gradient of the differences, respectively. The new gradient term in the above equation penalizes any sudden changes to the differences that occur horizontally or vertically. The loss does not consider any pixel that does not have ground-truth depth values in the scene. Considering only the valid pixels for $n$ ensures no complexities occur due to missing depth values near object boundaries or with peculiar surfaces while minimizing the loss function. Per-pixel depth loss suffers from slow convergence and yields unsatisfactory local solution [38].

*7.5.2 Re-projection-based losses.* The photometric similarity loss function is implicitly used to learn the depth. Several reconstruction losses were introduced over the years to measure the dissimilarity. In general, such a loss function is also named proxy loss [69]. Commonly, they use an appearance matching-based loss function to compare the dissimilarity of the two images during training. Two types of image reprojection losses can be identified as photometric loss [12, 43, 163, 210] or feature loss [194]. Photometric loss refers to the per-pixel color loss between the images where $I$ is the target image, $I'$ the synthesized image, and $pe$ the reprojection error,

$$SL_p = \sum pe(I, I').  \tag{10}$$

Later research emphasizes the importance of extending the loss function by making use of other factors such as disparity smoothness, consistency, and depth density. Despite that, the photometric reprojection loss has the disadvantage of converging to a local-minima during the self-supervised learning [164].

Some methods utilize the feature loss between the original and the synthesized image instead of relying only on pixel color value. Ye et al. [189] is an example where a feature-based warping loss for both single view depth estimation and visual odometry has been used.

*7.5.3 Ordinal regression-based losses.* Depth values carry strong ordinal correlation. Therefore, some methods utilize ordinal regression-based losses rather than simply modeling the task as a linear regression problem [23, 26, 33, 38]. Such methods discretize the depth values into a set of labels and learn to estimate depth as discrete values. For example, Fu et al. [38] use space-increasing discretization to produce an ordered set of labels for depth and train a network with an ordinal loss function. Several other methods also incorporate ordinal regression for depth estimation by considering the task as a classification task. Diaz et al. [33] propose a method that converts ground-truth depth data labels into soft probability distributions. They train the model to output an ordinal probability distribution that matches the ground truth via a categorical loss function.

## 8 GENERALIZATION AND DOMAIN ADAPTATION

In this section, we describe the methods that focus on developing deep learning models that generalize well to unseen data or new domains.

## 8.1 Generalization over unseen data

Generalization in terms of the deep learning model refers to the model's ability to estimate accurate results to data unseen during training. Existing methods utilize two strategies to achieve generalization: learning to generalize over a large distribution of data or adopting zero-shot and few-shot learning.

Several methods train the models to learn the general features over a large distribution of data. For instance, Bae et al. [6] propose to use shape-biased models such as transformer networks to achieve generalization. The study reveals that shape-biased models generalize well on out-of-distribution training datasets compared with texture-biased models such as CNNs. Yu et al. [192] propose to enhance generalization ability by combining knowledge distillation, intermediate prediction layers and loss re-balancing. Yang et al. [188] propose to decouple the camera parameters from pictorial cues to learn generalization over large data distribution. Ranftl et al. [126] achieve generalization using zero-shot cross dataset transfer. The method allows the model to train with multiple datasets. Therefore, the model learns from a large distribution of data.

Zero-shot and few-short learning methods estimate the depth of unseen data by training only on limited labeled data [47, 55]. They use auxiliary information such as textual descriptions or semantic embedding to achieve generalization.

Recently, visual foundation model-based methods show superior results in generalization [71, 183]. Such methods leverage limited labeled depth data along with large-scale unlabelled in-the-wild data for generalization. For example. DepthAnything [183] shows that powerful data augmentation strategies and pseudo annotations for large-scale data benefit deep-learning models to achieve excellent results in zero-shot depth estimation.

## 8.2 Domain adaptation

The monocular depth estimation problem expands over many domains, including indoor, outdoor, synthetic, real, day, night, urban, rural, and other environments. In the meantime, deep learning-based monocular depth estimation solutions suffer from domain shifting. In other words, the models perform poorly when the data distribution of the inference domain is different from the training data distribution. Due to the domain-biased training, the models cannot adapt to a new domain at inference. This also results in generalization glitches [77] for new domains. For practical scenarios, domain adaptation is crucial. Therefore, a considerable amount of research in the literature focus on solving this challenge [4, 48, 74, 93, 102, 131, 161, 204, 206, 207]. In this section, we describe two main approaches in the literature, for domain adaptation.

*8.2.1 Adaptation by fine tuning.* The idea is to initially train the models on one domain and then fine-tune them on the others. Roussel et al. [131] focus on the sensitivity of the pre-trained models to domain shift using two datasets and propose fine-tuning the pre-trained model with a new dataset to get better results. However, fine-tuning during inference is a burden for offline learning-based methods since it requires retraining every time for new inference. On the other hand, online fine-tuning [187, 200] learn to adapt to new domains as new data comes. Section 7.4 further describes these techniques. However, online adaptation leads to models that overfit the most recent data.

*8.2.2 Adaptation by data transformation.* These methods attempt to transform the data from one domain to match with the data of another domain [77]. They use a distribution alignment process from content to style image. The alignment minimizes the distance between the source and target feature distributions and style transfer [4]. Thus, the domains look similar. Atapour et al. [4] propose a GAN-based style transfer method to transform the style of synthetic data from a gaming environment to match the real data. Therefore, even though they train the model from synthetic data, the model produces more realistic depth maps for the real images during inference. The method uses adversarial learning to translate images from synthetic and real-world domains. AdaDepth [74] provides a GAN-based unsupervised approach for adapting from synthetic RGBD data to real scenes. Zhen et al. [207] also propose to use GANs to transform from synthetic to real. The method encourages the distribution mapping from synthetic to real but does not explicitly require the translated image to have a correct relationship to its corresponding depth. They adopt a wide-spectrum input translation network using adversarial loss for synthetic training and reconstruction loss for real image training. Zhao et al. [206] propose to perform real-to-synthetic translation by removing challenging regions from real images to bring them closer to the synthetic domain. However, they use a small amount of real data and a large amount of synthetic data.

Domain shift between day and night images is also a concern. Domain adaptation is necessary for the models trained with only day images to estimate accurate depth for night images. Liu et al. [93] propose to learn private (illumination) and invariant features (texture) separately. Invariant features are similar for the same scene day and night, while private features are not. They train a network to translate day images to night images while ensuring consistent invariant features.

## 9 DATASETS

Deep learning-based depth estimation models require RGB and depth data for training and evaluation. We briefly discuss the content of the selected datasets. Section A in the Appendix A further discusses the depth sensing techniques used in datasets and Section F provides short descriptions of the commonly used datasets.

### 9.1 Domain type

Generally, datasets include data from either indoor or outdoor environments. For instance, depth estimation for the autonomous driving task requires outdoor road datasets such as KITTI [105], Cityscapes [30], and Oxford robot car [101]. These include various data gathered for rural, highway, and urban areas with complex scenes containing pedestrians, road vehicles, and buildings during day and night. Indoor scenes are mostly living rooms and industrial halls under varying illuminations. Datasets containing animals, plants, and other nature-related scenes are infrequent. A variety of datasets covering many domains is essential for developing more generalized models.

### 9.2 Images and video types

Both natural and synthetic datasets are available. Synthetic data are existing digital 3D models (ShapeNet, Blendswap). Synthetic data provides accurate depth aligned with the objects in scenes. However, they lack realistic properties. Models trained with synthetic data fail to perform well for real images. Therefore, many studies have used real RGB-D datasets. We further divide real datasets into stereo or monocular datasets. We summarize 23 datasets in Table 1.

## 10 EVALUATION METRICS

Two types of metrics used for evaluation are the depth accuracy metric and the depth error metric. Depth accuracy metrics measure the accuracy as the percentage of pixels that exceed a predefined threshold error. Depth error metrics, on the other hand, provide the deviation of the estimated depth from the expected depth. A model performs better than others if the results for accuracy metrics are high or the results for error metrics are low. The evaluation metrics commonly used in the literature are summarized in Table 4 (Section G in Appendix A).

## 11 DISCUSSION AND FUTURE DIRECTIONS

Although existing methods show promising results in monocular depth estimation, there are still gaps that remain unresolved. We briefly outline those research gaps below.

- Inaccurate results for thin, small, and distant objects. The depth accuracy differs with the size of the object in the scene and its distance to the camera. Since thin and small objects such as plant leaves, animal fur, cat whiskers, bird feathers, and human hair only take a few pixels of the entire scene, estimating depth for such objects is challenging. Estimating depth with attention to thin and small objects is a potential direction for future research.

Table 1. A summary of 23 datasets in literature; "Cam. Intr": Camera Intrinsics, "Cam. Extr": Camera Extrinsics.

| Dataset | Year | Domain Type | Stereo /Mono | Depth Type | # scenes | Cam. Intr. | Cam. Extr. | Resolution | Purpose | Carrier |
|---|---|---|---|---|---|---|---|---|---|---|
| KITTI-360 [86] | 2022 | Outdoor | Stereo | LiDAR | 300K | Y | Y | 180° fish eye | Scene understanding, semantic SLAM | Vehicle |
| Tiktok Dataset [61] | 2022 | Indoor and Outdoor | Mono | - | 100K | Y | N | 1080×604 | social media videos of dressed humans, their foreground and uv coordinates | selfie videos |
| A2D2 [42] | 2020 | Outdoor | Stereo | LiDAR | 41277 | Y | Y | 1920×1208 | Auto. driving | Car |
| HR-VS [180] | 2019 | Synthetic, Outdoor | Stereo | Dense Eucl. | 780 | - | - | 2056×2464 | Auto. driving | Car |
| OmniHouse [166] | 2019 | Synthetic, fisheye | - | Dense | 2560 | - | - | 800×768 | Omni directional depth | - |
| Driving Stereo [181] | 2019 | Outdoor | Stereo | LiDAR | 182188 | Y | Y | 1762×800 | Auto. driving | Car |
| ApolloScape [56] | 2019 | Outdoor | Stereo | LiDAR | 5165 | Y | Y | 3130×960 | Auto. driving | Car |
| MegaDepth [85] | 2018 | real, internet images | Mono | Dense | 130K | N | N | 1600×1600 | 3D Reconstruction | - |
| TUM VI [139] | 2018 | Indoor, Outdoor | stereo | - | 28 | Y | Y | 1024×1024 | Odometry and SLAM | Aerial Vehicle |
| Oxford robot car [101] | 2017 | Outdoor | Stereo | LiDAR | 20M | Y | Y | 1280×960 | Auto. driving | Car |
| Blendswap [154] | 2017 | Synthetic, Indoor | - | - | 150 | - | - | 1024×436 | Optical flow eval. | - |
| ETH3D [138] | 2017 | Indoor, Outdoor | Stereo | LiDAR | 5 | Y | Y | 940×490 | 3D reconstruction | Person |
| SUNCG [143] | 2017 | Synthetic, Indoor | - | - | 45K | - | - | 640×480 | 3D reconstruction | - |
| Cityscapes [33] | 2016 | Outdoor | Stereo | - | 25K | Ego Motion | Ego Motion | 2048×1024 | segmentation , scene understanding | Car |
| EuroC [17] | 2016 | Indoor | Stereo | Leica MS50 laser | 11 | Y | Y | 752×480 | 3D reconstruction | Micro Aerial vehicle |
| SceneFlow [104] | 2016 | Synthetic | Stereo | - | 35k | - | - | 960×540 | Optical flow, 3D reconstruction | - |
| DTU [1] | 2016 | Real, small object | Multi view Stereo | Structured light scans | 80 | Y | Y | 1200×1600 | Small object reconstruction | Robot |
| KITTI [105] | 2015 | Outdoor | Stereo | LiDAR | 400 | Y | Y | 1242×376 | Autonomous driving | Car |
| Middlebury [137] | 2014 | Indoor | Stereo | Dense | 30 | Y | Y | 2948×1988 | Stereo matching | horizontal optical rail |
| SUN3D [169] | 2013 | Indoor | Mono | Xtion | 415 | - | Y | 640×480 | scene understanding | Person |
| NYU V2 [141] | 2012 | Indoor | Stereo | Kinect depth | 464 | N | N | 640×480 | Segmentation | Person |
| RGB-D SLAM [146] | 2012 | Indoor | Mono | Kinect depth | 39 | Y | Y | 640×480 | visual SLAM, odometry | Handheld, pioneer robot |
| Sintel [18] | 2012 | Synthetic | - | - | 1628 | - | - | 1024×436 | Optical flow eval. | - |

• Limitations in self-supervision. Existing self-supervised methods rely on matching pixels across multiple views. Matching pixels across frames is challenging in the presence of non-rigid motions of objects [114]. Current methods only attempt to mitigate the effect of rigid motions of objects using masking strategies. Stereomatching-based methods also suffer from stereo occlusions, a significant problem during training. Existing self-supervised methods produce outputs with texture copy artifacts and depth discontinuities in the presence of occlusions. Similar pixels in multiple regions cause ambiguity issues. Therefore, matching pixels across different views is challenging in similar regions due to complex lighting conditions, repetitive patterns, and texture-less regions [40].

• Inaccurate depth for complex materials. Existing methods fail to estimate accurate depth for complex materials with transparent and reflective surfaces (e.g., glass and mirrors in objects such as vehicles) as they lack context awareness. The shadows and illumination variances in inputs also cause inaccuracies.

• Methods limited to estimate at eye-level depth. Existing methods propose to estimate depth for views captured at eye level. However, estimating depth from an aerial view, wide-angle view (panoramic) and the point of view of a ground-level observer (worm's view) are crucial tasks that need further attention.

• Methods limited only for several applications. Most existing methods propose the general applications of monocular depth estimation, such as autonomous navigation and augmented reality. In contrast, only a few applications are available in medicine, biology and agriculture. For applications in the medical field, sub-pixel accuracy is essential. Existing methods cannot achieve that level of accuracy.

• Lack of explainability. Deep learning-based methods still suffer from the black-box nature. Due to that, the outputs are difficult to explain or interpret. Therefore, further research on improving the interpretability and explainability of the models without degrading their performance is another potential direction.

• Real-time depth estimation. Accurate depth estimation for real-time applications also remains a problem due to the availability of noisy feed and inconsistent lighting conditions in real-time inputs [72]. Performance limitation due to limited processing power is a significant factor causing them to perform poorly.

• CLIP models for depth. Contrastive Language-Image Pretraining (CLIP) has become an emerging trend in-depth estimation [5, 55, 142]. CLIP-based methods enable models to gain a vast general knowledge of estimating depth using the knowledge embedded with natural languages. However, CLIP models that are pretrained with human language prompts and human-set semantics are prone to human-biased choice of ordinal depth bins [5, 142]. Human bias limits the model's ability to generalize well and learn depth as an abstract concept. Therefore, further research on CLIP models for depth estimation is a potential direction.

Table 2 (Section C in Appendix A) provides a consolidated summary of the publications included in this survey.

## 12 CONCLUSION

This paper comprehensively surveys the deep learning-based monocular depth estimation methods. We discuss the challenges in estimating depth from monocular images and videos. We present a taxonomy and classify the methods based on their input/output modality, network architectures, and training supervision. We present a generic monocular depth estimation pipeline and a baseline encoder-decoder architecture, representing existing deep learning methods. Existing architectures have achieved high accuracy by incorporating multi-scale feature fusion techniques and attention mechanisms in their architectures. We identify increased attention toward domain adaptation and self-supervision techniques, considering the methods that emphasize training. Finally, we highlight the gaps identified from the survey, which are emerging topics for future research.

## REFERENCES

[1] Henrik Aanæs, Rasmus Ramsbøl Jensen, George Vogiatzis, Engin Tola, and Anders Bjorholm Dahl. 2016. Large-scale data for multiple-view stereopsis. *International Journal of Computer Vision* 120, 2 (2016), 153–168.

[2] Filippo Aleotti, Fabio Tosi, Matteo Poggi, and Stefano Mattoccia. 2018. Generative adversarial networks for unsupervised monocular depth prediction. In *Proceedings of the European conference on computer vision workshops*. 0–0.

[3] Ali Jahani Amiri, Shing Yan Loo, and Hong Zhang. 2019. Semi-supervised monocular depth estimation with left-right consistency using deep neural network. In *2019 IEEE International Conference on Robotics and Biomimetics (ROBIO)*. IEEE, 602–607.

[4] Amir Atapour-Abarghouei and Toby P Breckon. 2018. Real-time monocular depth estimation using synthetic data with domain adaptation via image style transfer. In *IEEE conference on computer vision and pattern recognition*. 2800–2810.

[5] Dylan Auty and Krystian Mikolajczyk. 2023. Learning to prompt clip for monocular depth estimation: Exploring the limits of human language. In *Proceedings of the IEEE/CVF International Conference on Computer Vision*. 2039–2047.

[6] Jinwoo Bae, Kyumin Hwang, and Sunghoon Im. 2023. A study on the generality of neural network structures for monocular depth estimation. *IEEE Transactions on Pattern Analysis and Machine Intelligence* (2023).

[7] Jinwoo Bae, Sungho Moon, and Sunghoon Im. 2022. Deep Digging into the Generalization of Self-supervised Monocular Depth Estimation. *arXiv preprint arXiv:2205.11083* (2022).

[8] Jinwoo Bae, Sungho Moon, and Sunghoon Im. 2022. MonoFormer: Towards Generalization of self-supervised monocular depth estimation with Transformers. *arXiv preprint arXiv:2205.11083* (2022).

[9] Jinwoo Bae, Sungho Moon, and Sunghoon Im. 2023. Deep digging into the generalization of self-supervised monocular depth estimation. In *Proceedings of the AAAI conference on artificial intelligence*, Vol. 37. 187–196.

[10] Jongbeom Baek, Gyeongnyeon Kim, and Seungryong Kim. 2022. Semi-Supervised Learning with Mutual Distillation for Monocular Depth Estimation. *arXiv preprint arXiv:2203.09737* (2022).

[11] Juan Luis Gonzalez Bello and Munchurl Kim. 2021. Self-Supervised Deep Monocular Depth Estimation with Ambiguity Boosting. *IEEE Transactions on Pattern Analysis and Machine Intelligence* (2021).

[12] Juan Luis Gonzalez Bello, Jaeho Moon, and Munchurl Kim. 2024. Self-Supervised Monocular Depth Estimation with Positional Shift Depth Variance and Adaptive Disparity Quantization. *IEEE Transactions on Image Processing* (2024).

[13] Shariq Farooq Bhat, Ibraheem Alhashim, and Peter Wonka. 2021. Adabins: Depth estimation using adaptive bins. In *IEEE conference on computer vision and pattern recognition*. 4009–4018.

[14] Amlaan Bhoi. 2019. Monocular depth estimation: A survey. *arXiv preprint arXiv:1901.09402* (2019).

[15] Jiawang Bian, Zhichao Li, Naiyan Wang, Huangying Zhan, Chunhua Shen, Ming-Ming Cheng, and Ian Reid. 2019. Unsupervised scale-consistent depth and ego-motion learning from monocular video. *Advances in neural information processing systems* 32 (2019).

[16] Aljaz Bozic, Pablo Palafox, Justus Thies, Angela Dai, and Matthias Nießner. 2021. Transformerfusion: Monocular rgb scene reconstruction using transformers. *Advances in neural information processing systems* 34 (2021).

[17] Michael Burri, Janosch Nikolic, Pascal Gohl, Thomas Schneider, Joern Rehder, Sammy Omari, Markus W Achtelik, and Roland Siegwart. 2016. The EuRoC micro aerial vehicle datasets. *The International Journal of Robotics Research* 35, 10 (2016), 1157–1163.

[18] Daniel J Butler, Jonas Wulff, Garrett B Stanley, and Michael J Black. 2012. A naturalistic open source movie for optical flow evaluation. In *European conference on computer vision*. Springer, 611–625.

[19] Yuanzhouhan Cao, Zifeng Wu, and Chunhua Shen. 2017. Estimating depth from monocular images as classification using deep fully convolutional residual networks. *IEEE Transactions on Circuits and Systems for Video Technology* 28, 11 (2017), 3174–3182.

[20] Vincent Casser, Soeren Pirk, Reza Mahjourian, and Anelia Angelova. 2019. Unsupervised monocular depth and ego-motion learning with structure and semantics. In *IEEE/CVF Conference on Computer Vision and Pattern Recognition Workshops*. 0–0.

[21] Po-Yi Chen, Alexander H Liu, Yen-Cheng Liu, and Yu-Chiang Frank Wang. 2019. Towards scene understanding: Unsupervised monocular depth estimation with semantic-aware representation. In *IEEE conference on computer vision and pattern recognition*. 2624–2632.

[22] Richard Chen, Faisal Mahmood, Alan Yuille, and Nicholas J Durr. 2018. Rethinking monocular depth estimation with adversarial training. *arXiv preprint arXiv:1808.07528* (2018).

[23] Tian Chen, Shijie An, Yuan Zhang, Chongyang Ma, Huayan Wang, Xiaoyan Guo, and Wen Zheng. 2020. Improving monocular depth estimation by leveraging structural awareness and complementary datasets. In *European conference on computer vision*. Springer, 90–108.

[24] Weifeng Chen, Zhao Fu, Dawei Yang, and Jia Deng. 2016. Single-image depth perception in the wild. *Advances in neural information processing systems* 29 (2016).

[25] Xiaotian Chen, Yuwang Wang, Xuejin Chen, and Wenjun Zeng. 2021. S2r-depthnet: Learning a generalizable depth-specific structural representation. In *IEEE/CVF Conference on Computer Vision and Pattern Recognition*. 3034–3043.

[26] Yuru Chen, Haitao Zhao, Zhengwei Hu, and Jingchao Peng. 2021. Attention-based context aggregation network for monocular depth estimation. *International Journal of Machine Learning and Cybernetics* 12, 6 (2021), 1583–1596.

[27] Zhi Chen, Xiaoqing Ye, Wei Yang, Zhenbo Xu, Xiao Tan, Zhikang Zou, Errui Ding, Xinming Zhang, and Liusheng Huang. 2021. Revealing the Reciprocal Relations Between Self-Supervised Stereo and Monocular Depth Estimation. In *Proceedings of the IEEE/CVF International Conference on Computer Vision*. 15529–15538.

[28] Zeyu Cheng, Yi Zhang, and Chengkai Tang. 2021. Swin-Depth: Using Transformers and Multi-Scale Fusion for Monocular-Based Depth Estimation. *IEEE Sensors Journal* 21, 23 (2021), 26912–26920.

[29] Jaehoon Cho, Dongbo Min, Youngjung Kim, and Kwanghoon Sohn. 2021. Deep monocular depth estimation leveraging a large-scale outdoor stereo dataset. *Expert Systems with Applications* 178 (2021), 114877.

[30] Marius Cordts, Mohamed Omran, Sebastian Ramos, Timo Rehfeld, Markus Enzweiler, Rodrigo Benenson, Uwe Franke, Stefan Roth, and Bernt Schiele. 2016. The cityscapes dataset for semantic urban scene understanding. In *IEEE conference on computer vision and pattern recognition*. 3213–3223.

[31] Arun CS Kumar, Suchendra M Bhandarkar, and Mukta Prasad. 2018. Depthnet: A recurrent neural network architecture for monocular depth prediction. In *IEEE conference on computer vision and pattern recognition Workshops*. 283–291.

[32] Arun CS Kumar, Suchendra M. Bhandarkar, and Mukta Prasad. 2018. Monocular Depth Prediction Using Generative Adversarial Networks. In *IEEE conference on computer vision and pattern recognition Workshops*.

[33] Raul Diaz and Amit Marathe. 2019. Soft labels for ordinal regression. In *Proceedings of the IEEE/CVF conference on computer vision and pattern recognition*. 4738–4747.

[34] Tom van Dijk and Guido de Croon. 2019. How do neural networks see depth in single images?. In *Proceedings of the IEEE/CVF International Conference on Computer Vision*. 2183–2191.

[35] David Eigen and Rob Fergus. 2015. Predicting depth, surface normals and semantic labels with a common multi-scale convolutional architecture. In *Proceedings of the IEEE international conference on computer vision*. 2650–2658.

[36] David Eigen, Christian Puhrsch, and Rob Fergus. 2014. Depth map prediction from a single image using a multi-scale deep network. *Advances in neural information processing systems* 27 (2014).

[37] Sara Elkerdawy, Hong Zhang, and Nilanjan Ray. 2019. Lightweight monocular depth estimation model by joint end-to-end filter pruning. In *2019 IEEE International Conference on Image Processing (ICIP)*. IEEE, 4290–4294.

[38] Huan Fu, Mingming Gong, Chaohui Wang, Kayhan Batmanghelich, and Dacheng Tao. 2018. Deep ordinal regression network for monocular depth estimation. In *IEEE conference on computer vision and pattern recognition*. 2002–2011.

[39] Ravi Garg, Vijay Kumar Bg, Gustavo Carneiro, and Ian Reid. 2016. Unsupervised cnn for single view depth estimation: Geometry to the rescue. In *European conference on computer vision*. Springer, 740–756.

[40] Stefano Gasperini, Nils Morbitzer, HyunJun Jung, Nassir Navab, and Federico Tombari. 2023. Robust monocular depth estimation under challenging conditions. In *Proceedings of the IEEE/CVF international conference on computer vision*. 8177–8186.

[41] Daniel Gehrig, Michelle Rüegg, Mathias Gehrig, Javier Hidalgo-Carrió, and Davide Scaramuzza. 2021. Combining events and frames using recurrent asynchronous multimodal networks for monocular depth prediction. *IEEE Robotics and Automation Letters* 6, 2 (2021), 2822–2829.

[42] Jakob Geyer, Yohannes Kassahun, Mentar Mahmudi, Xavier Ricou, Rupesh Durgesh, Andrew S Chung, Lorenz Hauswald, Viet Hoang Pham, Maximilian Mühlegg, Sebastian Dorn, et al. 2020. A2d2: Audi autonomous driving dataset. *arXiv preprint arXiv:2004.06320* (2020).

[43] Clément Godard, Oisin Mac Aodha, and Gabriel J Brostow. 2017. Unsupervised monocular depth estimation with left-right consistency. In *IEEE conference on computer vision and pattern recognition*. 270–279.

[44] Clément Godard, Oisin Mac Aodha, Michael Firman, and Gabriel J Brostow. 2019. Digging into self-supervised monocular depth estimation. In *Proceedings of the IEEE International Conference on Computer Vision*. 3828–3838.

[45] Vitor Guizilini, Rares Ambrus, Sudeep Pillai, Allan Raventos, and Adrien Gaidon. 2020. 3d packing for self-supervised monocular depth estimation. In *IEEE conference on computer vision and pattern recognition*. 2485–2494.

[46] Vitor Guizilini, Jie Li, Rares Ambrus, Sudeep Pillai, and Adrien Gaidon. 2020. Robust semi-supervised monocular depth estimation with reprojected distances. In *Conference on robot learning*. PMLR, 503–512.

[47] Vitor Guizilini, Igor Vasiljevic, Dian Chen, Rares. Ambrus., and Adrien Gaidon. 2023. Towards zero-shot scale-aware monocular depth estimation. In *Proceedings of the IEEE/CVF International Conference on Computer Vision*. 9233–9243.

[48] Rui Guo, Babajide Ayinde, Hao Sun, Haritha Muralidharan, and Kentaro Oguchi. 2019. Monocular depth estimation using synthetic images with shadow removal. In *2019 IEEE Intelligent Transportation Systems Conference (ITSC)*. IEEE, 1432–1439.

[49] Xian-Feng Han, Hamid Laga, and Mohammed Bennamoun. 2019. Image-based 3D object reconstruction: State-of-the-art and trends in the deep learning era. *IEEE transactions on pattern analysis and machine intelligence* 43, 5 (2019), 1578–1604.

[50] Zhixiang Hao, Yu Li, Shaodi You, and Feng Lu. 2018. Detail preserving depth estimation from a single image using attention guided networks. In *2018 international conference on 3D vision (3DV)*. IEEE, 304–313.

[51] Lei He, Jiwen Lu, Guanghui Wang, Shiyu Song, and Jie Zhou. 2021. SOSD-Net: Joint semantic object segmentation and depth estimation from monocular images. *Neurocomputing* 440 (2021), 251–263.

[52] Mu He, Le Hui, Yikai Bian, Jian Ren, Jin Xie, and Jian Yang. 2022. RA-Depth: Resolution Adaptive Self-supervised Monocular Depth Estimation. In *European Conference on Computer Vision*. Springer, 565–581.

[53] Minhyeok Heo, Jaehan Lee, Kyung-Rae Kim, Han-Ul Kim, and Chang-Su Kim. 2018. Monocular depth estimation using whole strip masking and reliability-based refinement. In *Proceedings of the European Conference on Computer Vision (ECCV)*. 36–51.

[54] Junjie Hu, Yan Zhang, and Takayuki Okatani. 2019. Visualization of convolutional neural networks for monocular depth estimation. In *Proceedings of the IEEE/CVF International Conference on Computer Vision*. 3869–3878.

[55] Xueting Hu, Ce Zhang, Yi Zhang, Bowen Hai, Ke Yu, and Zhihai He. 2024. Learning to adapt CLIP for few-shot monocular depth estimation. In *Proceedings of the IEEE/CVF Winter Conference on Applications of Computer Vision*. 5594–5603.

[56] Xinyu Huang, Peng Wang, Xinjing Cheng, Dingfu Zhou, Qichuan Geng, and Ruigang Yang. 2019. The apolloscape open dataset for autonomous driving and its application. *IEEE transactions on pattern analysis and machine intelligence* 42, 10 (2019), 2702–2719.

[57] Zixuan Huang, Junming Fan, Shenggan Cheng, Shuai Yi, Xiaogang Wang, and Hongsheng Li. 2019. Hms-net: Hierarchical multi-scale sparsity-invariant network for sparse depth completion. *IEEE Transactions on Image Processing* 29 (2019), 3429–3441.

[58] Tak-Wai Hui. 2022. RM-Depth: Unsupervised Learning of Recurrent Monocular Depth in Dynamic Scenes. In *IEEE/CVF Conference on Computer Vision and Pattern Recognition*. 1675–1684.

[59] Lam Huynh, Phong Nguyen-Ha, Jiri Matas, Esa Rahtu, and Janne Heikkilä. 2020. Guiding monocular depth estimation using depth-attention volume. In *European Conference on Computer Vision*. Springer, 581–597.

[60] Yasamin Jafarian and Hyun Soo Park. 2021. Learning high fidelity depths of dressed humans by watching social media dance videos. In *IEEE/CVF Conference on Computer Vision and Pattern Recognition*. 12753–12762.

[61] Yasamin Jafarian and Hyun Soo Park. 2022. Self-supervised 3D Representation Learning of Dressed Humans from Social Media Videos. *IEEE Transactions on Pattern Analysis and Machine Intelligence* (2022).

[62] Pan Ji, Runze Li, Bir Bhanu, and Yi Xu. 2021. Monoindoor: Towards good practice of self-supervised monocular depth estimation for indoor environments. In *IEEE conference on computer vision and pattern recognition*. 12787–12796.

[63] Rongrong Ji, Ke Li, Yan Wang, Xiaoshuai Sun, Feng Guo, Xiaowei Guo, Yongjian Wu, Feiyue Huang, and Jiebo Luo. 2019. Semi-supervised adversarial monocular depth estimation. *IEEE transactions on pattern analysis and machine intelligence* 42, 10 (2019), 2410–2422.

[64] Jianbo Jiao, Ying Cao, Yibing Song, and Rynson Lau. 2018. Look deeper into depth: Monocular depth estimation with semantic booster and attention-driven loss. In *Proceedings of the European conference on computer vision (ECCV)*. 53–69.

[65] Adrian Johnston and Gustavo Carneiro. 2020. Self-supervised monocular trained depth estimation using self-attention and discrete disparity volume. In *IEEE conference on computer vision and pattern recognition*. 4756–4765.

[66] Hyungjoo Jung, Youngjung Kim, Dongbo Min, Changjae Oh, and Kwanghoon Sohn. 2017. Depth prediction from a single image with conditional adversarial networks. In *2017 IEEE International Conference on Image Processing (ICIP)*. IEEE, 1717–1721.

[67] Kevin Karsch, Ce Liu, and Sing Bing Kang. 2014. Depth transfer: Depth extraction from video using non-parametric sampling. *IEEE transactions on pattern analysis and machine intelligence* 36, 11 (2014), 2144–2158.

[68] Salman Khan, Muzammal Naseer, Munawar Hayat, Syed Waqas Zamir, Fahad Shahbaz Khan, and Mubarak Shah. 2021. Transformers in vision: A survey. *ACM Computing Surveys (CSUR)* (2021).

[69] Sungyeon Kim, Dongwon Kim, Minsu Cho, and Suha Kwak. 2020. Proxy Anchor Loss for Deep Metric Learning. In *2020 IEEE conference on computer vision and pattern recognition*. 3235–3244. https://doi.org/10.1109/CVPR42600.2020.00330

[70] Youngjung Kim, Hyungjoo Jung, Dongbo Min, and Kwanghoon Sohn. 2018. Deep monocular depth estimation via integration of global and local predictions. *IEEE transactions on Image Processing* 27, 8 (2018), 4131–4144.

[71] Alexander Kirillov, Eric Mintun, Nikhila Ravi, Hanzi Mao, Chloe Rolland, Laura Gustafson, Tete Xiao, Spencer Whitehead, Alexander C Berg, Wan-Yen Lo, et al. 2023. Segment anything. In *Proceedings of the IEEE/CVF International Conference on Computer Vision*. 4015–4026.

[72] Johannes Kopf, Xuejian Rong, and Jia Bin Huang. 2021. Robust Consistent Video Depth Estimation. *Proceedings of the IEEE computer society conference on computer vision and pattern recognition*. https://doi.org/10.1109/CVPR46437.2021.00166

[73] Dimce Kostadinov and Zoran Ivanovski. 2012. Single image depth estimation using local gradient-based features. In *2012 19th International Conference on Systems, Signals and Image Processing (IWSSIP)*. IEEE, 596–599.

[74] Jogendra Nath Kundu, Phani Krishna Uppala, Anuj Pahuja, and R Venkatesh Babu. 2018. Adadepth: Unsupervised content congruent adaptation for depth estimation. In *IEEE conference on computer vision and pattern recognition*. 2656–2665.

[75] Yevhen Kuznietsov, Marc Proesmans, and Luc Van Gool. 2021. Comoda: Continuous monocular depth adaptation using past experiences. In *Proceedings of the IEEE/CVF Winter Conference on Applications of Computer Vision*. 2907–2917.

[76] Yevhen Kuznietsov, Jorg Stuckler, and Bastian Leibe. 2017. Semi-supervised deep learning for monocular depth map prediction. In *IEEE conference on computer vision and pattern recognition*. 6647–6655.

[77] Hamid Laga, Laurent Valentin Jospin, Farid Boussaid, and Mohammed Bennamoun. 2020. A survey on deep learning techniques for stereo-based depth estimation. *IEEE transactions on pattern analysis and machine intelligence* 44, 4 (2020), 1738–1764.

[78] Iro Laina, Christian Rupprecht, Vasileios Belagiannis, Federico Tombari, and Nassir Navab. 2016. Deeper depth prediction with fully convolutional residual networks. In *2016 Fourth international conference on 3D vision (3DV)*. IEEE, 239–248.

[79] Jin Han Lee, Myung-Kyu Han, Dong Wook Ko, and Il Hong Suh. 2019. From big to small: Multi-scale local planar guidance for monocular depth estimation. *arXiv preprint arXiv:1907.10326* (2019).

[80] Jae-Han Lee, Minhyeok Heo, Kyung-Rae Kim, and Chang-Su Kim. 2018. Single-image depth estimation based on fourier domain analysis. In *IEEE conference on computer vision and pattern recognition*. 330–339.

[81] Minhyeok Lee, Sangwon Hwang, Chaewon Park, and Sangyoun Lee. 2022. EdgeConv with Attention Module for Monocular Depth Estimation. In *Proceedings of the IEEE/CVF Winter Conference on Applications of Computer Vision*. 2858–2867.

[82] Zeyu Lei, Yan Wang, Zijian Li, and Junyao Yang. 2021. Attention based multilayer feature fusion convolutional neural network for unsupervised monocular depth estimation. *Neurocomputing* 423 (2021), 343–352.

[83] Bo Li, Chunhua Shen, Yuchao Dai, Anton Van Den Hengel, and Mingyi He. 2015. Depth and surface normal estimation from monocular images using regression on deep features and hierarchical crfs. In *IEEE conference on computer vision and pattern recognition*. 1119–1127.

[84] Ruibo Li, Ke Xian, Chunhua Shen, Zhiguo Cao, Hao Lu, and Lingxiao Hang. 2018. Deep attention-based classification network for robust depth prediction. In *Asian Conference on Computer Vision*. Springer, 663–678.

[85] Zhengqi Li and Noah Snavely. 2018. Megadepth: Learning single-view depth prediction from internet photos. In *IEEE conference on computer vision and pattern recognition*. 2041–2050.

[86] Yiyi Liao, Jun Xie, and Andreas Geiger. 2022. KITTI-360: A novel dataset and benchmarks for urban scene understanding in 2d and 3d. *IEEE Transactions on Pattern Analysis and Machine Intelligence* (2022).

[87] Julian Lienen, Eyke Hullermeier, Ralph Ewerth, and Nils Nommensen. 2021. Monocular depth estimation via listwise ranking using the plackett-luce model. In *IEEE/CVF Conference on Computer Vision and Pattern Recognition*. 14595–14604.

[88] Jingyu Lin, Xiangyang Ji, Wenli Xu, and Qionghai Dai. 2013. Absolute depth estimation from a single defocused image. *IEEE Transactions on Image Processing* 22, 11 (2013), 4545–4550.

[89] Chuanwu Ling, Xiaogang Zhang, and Hua Chen. 2021. Unsupervised monocular depth estimation using attention and multi-warp reconstruction. *IEEE Transactions on Multimedia* (2021).

[90] Fayao Liu, Chunhua Shen, and Guosheng Lin. 2015. Deep convolutional neural fields for depth estimation from a single image. In *IEEE conference on computer vision and pattern recognition*. 5162–5170.

[91] Fayao Liu, Chunhua Shen, Guosheng Lin, and Ian Reid. 2015. Learning depth from single monocular images using deep convolutional neural fields. *IEEE transactions on pattern analysis and machine intelligence* 38, 10 (2015), 2024–2039.

[92] Jing Liu, Xiaona Zhang, Zhaoxin Li, and Tianlu Mao. 2021. Multi-scale residual pyramid attention network for monocular depth estimation. In *2020 25th International Conference on Pattern Recognition (ICPR)*. IEEE, 5137–5144.

[93] Lina Liu, Xibin Song, Mengmeng Wang, Yong Liu, and Liangjun Zhang. 2021. Self-supervised monocular depth estimation for all day images using domain separation. In *Proceedings of the IEEE/CVF International Conference on Computer Vision*. 12737–12746.

[94] Miaomiao Liu, Mathieu Salzmann, and Xuming He. 2014. Discrete-continuous depth estimation from a single image. In *IEEE conference on computer vision and pattern recognition*. 716–723.

[95] Peng Liu, Zonghua Zhang, Zhaozong Meng, and Nan Gao. 2020. Joint attention mechanisms for monocular depth estimation with multi-scale convolutions and adaptive weight adjustment. *IEEE Access* 8 (2020), 184437–184450.

[96] Wu Liu and Tao Mei. 2022. Recent Advances of Monocular 2D and 3D Human Pose Estimation: A Deep Learning Perspective. *ACM Computing Surveys (CSUR)* (2022).

[97] Shing Yan Loo, Moein Shakeri, Sai Hong Tang, Syamsiah Mashohor, and Hong Zhang. 2021. Online Mutual Adaptation of Deep Depth Prediction and Visual SLAM. *CoRR* (2021).

[98] Ivan Lopes, Tuan-Hung Vu, and Raoul de Charette. 2023. Cross-task attention mechanism for dense multi-task learning. In *Proceedings of the IEEE/CVF Winter Conference on Applications of Computer Vision*. 2329–2338.

[99] Xuan Luo, Jia-Bin Huang, Richard Szeliski, Kevin Matzen, and Johannes Kopf. 2020. Consistent video depth estimation. *ACM Transactions on Graphics (ToG)* 39, 4 (2020), 71–1.

[100] Poggi M. 2020. *Learning and Understanding Single Image Depth Estimation in the Wild*. https://drive.google.com/file/d/17Bzlj_KZTXD_WheehKNup7f9BY5_BdHK/view

[101] Will Maddern, Geoffrey Pascoe, Chris Linegar, and Paul Newman. 2017. 1 year, 1000 km: The Oxford RobotCar dataset. *The International Journal of Robotics Research* 36, 1 (2017), 3–15.

[102] Michele Mancini, Gabriele Costante, Paolo Valigi, Thomas A Ciarfuglia, Jeffrey Delmerico, and Davide Scaramuzza. 2017. Toward domain independence for learning-based monocular depth estimation. *IEEE Robotics and Automation Letters* 2, 3 (2017), 1778–1785.

[103] Dmitrii Maslov and Ilya Makarov. 2020. Online supervised attention-based recurrent depth estimation from monocular video. *PeerJ Computer Science* 6 (2020), e317.

[104] Nikolaus Mayer, Eddy Ilg, Philip Hausser, Philipp Fischer, Daniel Cremers, Alexey Dosovitskiy, and Thomas Brox. 2016. A large dataset to train convolutional networks for disparity, optical flow, and scene flow estimation. In *IEEE conference on computer vision and pattern recognition*. 4040–4048.

[105] Moritz Menze and Andreas Geiger. 2015. Object scene flow for autonomous vehicles. In *IEEE conference on computer vision and pattern recognition*. 3061–3070.

[106] S. H.Mahdi Miangoleh, Sebastian Dille, Long Mai, Sylvain Paris, and Yağız Aksoy. 2021. Boosting Monocular Depth Estimation Models to High-resolution via Content-adaptive Multi-Resolution Merging. *Proceedings of the IEEE computer society conference on computer vision and pattern recognition*. https://doi.org/10.1109/CVPR46437.2021.00956

[107] Yue Ming, Xuyang Meng, Chunxiao Fan, and Hui Yu. 2021. Deep learning for monocular depth estimation: A review. *Neurocomputing* 438 (2021), 14–33.

[108] Lawrence Mutimbu and Antonio Robles-Kelly. 2013. A relaxed factorial Markov random field for colour and depth estimation from a single foggy image. In *2013 IEEE International Conference on Image Processing*. IEEE, 355–359.

[109] Taher Naderi, Amir Sadovnik, Jason Hayward, and Hairong Qi. 2022. Monocular Depth Estimation with Adaptive Geometric Attention. In *Proceedings of the IEEE/CVF Winter Conference on Applications of Computer Vision*. 944–954.

[110] Muhammad Muzammal Naseer, Kanchana Ranasinghe, Salman H Khan, Munawar Hayat, Fahad Shahbaz Khan, and Ming-Hsuan Yang. 2021. Intriguing properties of vision transformers. *Advances in Neural Information Processing Systems* 34 (2021), 23296–23308.

[111] Zhaoyang Niu, Guoqiang Zhong, and Hui Yu. 2021. A review on the attention mechanism of deep learning. *Neurocomputing* 452 (2021), 48–62.

[112] Vaishakh Patil, Wouter Van Gansbeke, Dengxin Dai, and Luc Van Gool. 2020. Don't forget the past: Recurrent depth estimation from monocular video. *IEEE Robotics and Automation Letters* 5, 4 (2020), 6813–6820.

[113] Rui Peng, Ronggang Wang, Yawen Lai, Luyang Tang, and Yangang Cai. 2021. Excavating the Potential Capacity of Self-Supervised Monocular Depth Estimation. In *IEEE conference on computer vision and pattern recognition*. 15560–15569.

[114] Andra Petrovai and Sergiu Nedevschi. 2022. Exploiting Pseudo Labels in a Self-Supervised Learning Framework for Improved Monocular Depth Estimation. In *IEEE/CVF Conference on Computer Vision and Pattern Recognition*. 1578–1588.

[115] Luigi Piccinelli, Christos Sakaridis, and Fisher Yu. 2023. iDisc: Internal discretization for monocular depth estimation. In *Proceedings of the IEEE/CVF Conference on Computer Vision and Pattern Recognition*. 21477–21487.

[116] Sudeep Pillai, Rareş Ambruş, and Adrien Gaidon. 2019. SuperDepth: Self-Supervised, Super-Resolved Monocular Depth Estimation. In *2019 International Conference on Robotics and Automation (ICRA)*. 9250–9256. https://doi.org/10.1109/ICRA.2019.8793621

[117] Andrea Pilzer, Dan Xu, Mihai Puscas, Elisa Ricci, and Nicu Sebe. 2018. Unsupervised adversarial depth estimation using cycled generative networks. In *2018 international conference on 3D vision (3DV)*. IEEE, 587–595.

[118] Giovanni Pintore, Marco Agus, Eva Almansa, Jens Schneider, and Enrico Gobbetti. 2021. SliceNet: deep dense depth estimation from a single indoor panorama using a slice-based representation. In *IEEE conference on computer vision and pattern recognition*. 11536–11545.

[119] Matteo Poggi, Fabio Tosi, Konstantinos Batsos, Philippos Mordohai, and Stefano Mattoccia. 2021. On the synergies between machine learning and binocular stereo for depth estimation from images: a survey. *IEEE trans. on pattern anal. and machine intelligence* (2021).

[120] Matteo Poggi, Fabio Tosi, and Stefano Mattoccia. 2018. Learning monocular depth estimation with unsupervised trinocular assumptions. In *2018 international conference on 3D vision (3DV)*. IEEE, 324–333.

[121] Mahsa T Pourazad, Panos Nasiopoulos, and Ali Bashashati. 2010. Random forests-based 2D-to-3D video conversion. In *2010 17th IEEE International Conference on Electronics, Circuits and Systems*. IEEE, 150–153.

[122] Mihai Marian Puscas, Dan Xu, Andrea Pilzer, and Niculae Sebe. 2019. Structured coupled generative adversarial networks for unsupervised monocular depth estimation. In *2019 international conference on 3D vision (3DV)*. IEEE, 18–26.

[123] Zequn Qin and Xi Li. 2022. MonoGround: Detecting Monocular 3D Objects From the Ground. In *IEEE/CVF Conference on Computer Vision and Pattern Recognition*. 3793–3802.

[124] Michael Ramamonjisoa and Vincent Lepetit. 2019. Sharpnet: Fast and accurate recovery of occluding contours in monocular depth estimation. In *Proceedings of the IEEE/CVF International Conference on Computer Vision Workshops*. 0–0.

[125] René Ranftl, Alexey Bochkovskiy, and Vladlen Koltun. 2021. Vision transformers for dense prediction. In *IEEE conference on computer vision and pattern recognition*. 12179–12188.

[126] René Ranftl, Katrin Lasinger, David Hafner, Konrad Schindler, and Vladlen Koltun. 2020. Towards robust monocular depth estimation: Mixing datasets for zero-shot cross-dataset transfer. *IEEE transactions on pattern analysis and machine intelligence* 44, 3 (2020), 1623–1637.

[127] Rene Ranftl, Vibhav Vineet, Qifeng Chen, and Vladlen Koltun. 2016. Dense monocular depth estimation in complex dynamic scenes. In *IEEE conference on computer vision and pattern recognition*. 4058–4066.

[128] Haoyu Ren, Aman Raj, Mostafa El-Khamy, and Jungwon Lee. 2020. Suw-learn: Joint supervised, unsupervised, weakly supervised deep learning for monocular depth estimation. In *IEEE conference on computer vision and pattern recognition Workshops*. 750–751.

[129] Vamshi Krishna Repala and Shiv Ram Dubey. 2019. Dual CNN models for unsupervised monocular depth estimation. In *International Conference on Pattern Recognition and Machine Intelligence*. Springer, 209–217.

[130] Manuel Rey-Area, Mingze Yuan, and Christian Richardt. 2022. 360monodepth: High-resolution 360deg monocular depth estimation. In *Proceedings of the IEEE/CVF Conference on Computer Vision and Pattern Recognition*. 3762–3772.

[131] Tom Roussel, Luc Van Eycken, and Tinne Tuytelaars. 2019. Monocular depth estimation in new environments with absolute scale. In *2019 IEEE/RSJ International Conference on Intelligent Robots and Systems (IROS)*. IEEE, 1735–1741.

[132] Anirban Roy and Sinisa Todorovic. 2016. Monocular depth estimation using neural regression forest. In *IEEE conference on computer vision and pattern recognition*. 5506–5514.

[133] Abhinav Sagar. 2022. Monocular depth estimation using multi scale neural network and feature fusion. In *Proceedings of the IEEE/CVF Winter Conference on Applications of Computer Vision*. 656–662.

[134] Ashutosh Saxena, Sung Chung, and Andrew Ng. 2005. Learning Depth from Single Monocular Images. In *Advances in neural information processing systems*, Y. Weiss, B. Schölkopf, and J. Platt (Eds.), Vol. 18. MIT Press.

[135] Ashutosh Saxena, Min Sun, and Andrew Y Ng. 2008. Make3d: Learning 3d scene structure from a single still image. *IEEE transactions on pattern analysis and machine intelligence* 31, 5 (2008), 824–840.

[136] Saurabh Saxena, Charles Herrmann, Junhwa Hur, Abhishek Kar, Mohammad Norouzi, Deqing Sun, and David J Fleet. 2024. The surprising effectiveness of diffusion models for optical flow and monocular depth estimation. *Advances in Neural Information Processing Systems* 36 (2024).

[137] Daniel Scharstein, Heiko Hirschmüller, York Kitajima, Greg Krathwohl, Nera Nešić, Xi Wang, and Porter Westling. 2014. High-resolution stereo datasets with subpixel-accurate ground truth. In *German conference on pattern recognition*. Springer, 31–42.

[138] Thomas Schops, Johannes L Schonberger, Silvano Galliani, Torsten Sattler, Konrad Schindler, Marc Pollefeys, and Andreas Geiger. 2017. A multi-view stereo benchmark with high-resolution images and multi-camera videos. In *IEEE conference on computer vision and pattern recognition*. 3260–3269.

[139] David Schubert, Thore Goll, Nikolaus Demmel, Vladyslav Usenko, Jörg Stückler, and Daniel Cremers. 2018. The TUM VI benchmark for evaluating visual-inertial odometry. In *2018 IEEE/RSJ International Conference on Intelligent Robots and Systems (IROS)*. IEEE, 1680–1687.

[140] Shuwei Shao, Zhongcai Pei, Xingming Wu, Zhong Liu, Weihai Chen, and Zhengguo Li. 2024. Iebins: Iterative elastic bins for monocular depth estimation. *Advances in Neural Information Processing Systems* 36 (2024).

[141] Nathan Silberman, Derek Hoiem, Pushmeet Kohli, and Rob Fergus. 2012. Indoor segmentation and support inference from rgbd images. In *European conference on computer vision*. Springer, 746–760.

[142] Eunjin Son and Sang Jun Lee. 2024. CaBins: CLIP-based Adaptive Bins for Monocular Depth Estimation. In *Proceedings of the IEEE/CVF Conference on Computer Vision and Pattern Recognition*. 4557–4567.

[143] Shuran Song, Fisher Yu, Andy Zeng, Angel X Chang, Manolis Savva, and Thomas Funkhouser. 2017. Semantic scene completion from a single depth image. In *IEEE conference on computer vision and pattern recognition*. 1746–1754.

[144] Wenfeng Song, Shuai Li, Ji Liu, Aimin Hao, Qinping Zhao, and Hong Qin. 2019. Contextualized CNN for scene-aware depth estimation from single RGB image. *IEEE Transactions on Multimedia* 22, 5 (2019), 1220–1233.

[145] Xibin Song, Wei Li, Dingfu Zhou, Yuchao Dai, Jin Fang, Hongdong Li, and Liangjun Zhang. 2021. MLDA-Net: Multi-level dual attention-based network for self-supervised monocular depth estimation. *IEEE Transactions on Image Processing* 30 (2021), 4691–4705.

[146] Jürgen Sturm, Nikolas Engelhard, Felix Endres, Wolfram Burgard, and Daniel Cremers. 2012. A benchmark for the evaluation of RGB-D SLAM systems. In *IEEE/RSJ international conference on intelligent robots and systems*. IEEE, 573–580.

[147] Wen Su and Haifeng Zhang. 2020. Soft regression of monocular depth using scale-semantic exchange network. *IEEE Access* 8 (2020), 114930–114939.

[148] Wen Su, Haifeng Zhang, Quan Zhou, Wenzhen Yang, and Zengfu Wang. 2020. Monocular depth estimation using information exchange network. *IEEE Transactions on Intelligent Transportation Systems* 22, 6 (2020), 3491–3503.

[149] Libo Sun, Jia-Wang Bian, Huangying Zhan, Wei Yin, Ian Reid, and Chunhua Shen. 2023. Sc-depthv3: Robust self-supervised monocular depth estimation for dynamic scenes. *IEEE Transactions on Pattern Analysis and Machine Intelligence* (2023).

[150] K Swaraja, V Akshitha, K Pranav, B Vyshnavi, V Sai Akhil, K Meenakshi, Padmavathi Kora, Himabindu Valiveti, and Chaitanya Duggineni. 2021. Monocular Depth Estimation using Transfer learning-An Overview. In *E3S Web of Conferences*, Vol. 309. EDP Sciences.

[151] Dong Tian, Po-Lin Lai, Patrick Lopez, and Cristina Gomila. 2009. View synthesis techniques for 3D video. In *Applications of Digital Image Processing XXXII*, Vol. 7443. SPIE, 233–243.

[152] Hu Tian and Fei Li. 2019. Semi-supervised depth estimation from a single image based on confidence learning. In *ICASSP 2019-2019 IEEE International Conference on Acoustics, Speech and Signal Processing (ICASSP)*. IEEE, 8573–8577.

[153] Fabio Tosi, Filippo Aleotti, Matteo Poggi, and Stefano Mattoccia. 2019. Learning monocular depth estimation infusing traditional stereo knowledge. In *IEEE conference on computer vision and pattern recognition*. 9799–9809.

[154] Benjamin Ummenhofer, Huizhong Zhou, Jonas Uhrig, Nikolaus Mayer, Eddy Ilg, Alexey Dosovitskiy, and Thomas Brox. 2017. Demon: Depth and motion network for learning monocular stereo. In *IEEE conference on computer vision and pattern recognition*. 5038–5047.

[155] Yannick Verdié, Jifei Song, Barnabé Mas, Benjamin Busam, Ales Leonardis, and Steven McDonagh. 2022. CroMo: Cross-Modal Learning for Monocular Depth Estimation. In *IEEE/CVF Conference on Computer Vision and Pattern Recognition*. 3937–3947.

[156] Pulkit Vyas, Chirag Saxena, Anwesh Badapanda, and Anurag Goswami. 2022. Outdoor Monocular Depth Estimation: A Research Review. *arXiv preprint arXiv:2205.01399* (2022).

[157] Brandon Wagstaff and Jonathan Kelly. 2021. Self-supervised scale recovery for monocular depth and egomotion estimation. In *2021 IEEE/RSJ International Conference on Intelligent Robots and Systems (IROS)*. IEEE, 2620–2627.

[158] Matthew Wallingford, Hao Li, Alessandro Achille, Avinash Ravichandran, Charless Fowlkes, Rahul Bhotika, and Stefano Soatto. 2022. Task adaptive parameter sharing for multi-task learning. In *Proceedings of the IEEE/CVF Conference on Computer Vision and Pattern Recognition*. 7561–7570.

[159] Guangming Wang, Chi Zhang, Hesheng Wang, Jingchuan Wang, Yong Wang, and Xinlei Wang. 2020. Unsupervised learning of depth, optical flow and pose with occlusion from 3d geometry. *IEEE Transactions on Intelligent Transportation Systems* 23, 1 (2020), 308–320.

[160] Lijun Wang, Jianming Zhang, Oliver Wang, Zhe Lin, and Huchuan Lu. 2020. Sdc-depth: Semantic divide-and-conquer network for monocular depth estimation. In *IEEE conference on computer vision and pattern recognition*. 541–550.

[161] Qin Wang, Dengxin Dai, Lukas Hoyer, Luc Van Gool, and Olga Fink. 2021. Domain adaptive semantic segmentation with self-supervised depth estimation. In *Proceedings of the IEEE/CVF International Conference on Computer Vision*. 8515–8525.

[162] Rui Wang, Stephen M Pizer, and Jan-Michael Frahm. 2019. Recurrent neural network for (un-) supervised learning of monocular video visual odometry and depth. In *IEEE conference on computer vision and pattern recognition*. 5555–5564.

[163] Ruoyu Wang, Zehao Yu, and Shenghua Gao. 2023. PlaneDepth: Self-supervised depth estimation via orthogonal planes. In *Proceedings of the IEEE/CVF Conference on Computer Vision and Pattern Recognition*. 21425–21434.

[164] Jamie Watson, Michael Firman, Gabriel J Brostow, and Daniyar Turmukhambetov. 2019. Self-supervised monocular depth hints. In *IEEE conference on computer vision and pattern recognition*. 2162–2171.

[165] Jamie Watson, Oisin Mac Aodha, Victor Prisacariu, Gabriel Brostow, and Michael Firman. 2021. The temporal opportunist: Self-supervised multi-frame monocular depth. In *IEEE conference on computer vision and pattern recognition*. 1164–1174.

[166] Changhee Won, Jongbin Ryu, and Jongwoo Lim. 2019. Omnimvs: End-to-end learning for omnidirectional stereo matching. In *Proceedings of the IEEE/CVF International Conference on Computer Vision*. 8987–8996.

[167] Sanghyun Woo, Jongchan Park, Joon-Young Lee, and In So Kweon. 2018. Cbam: Convolutional block attention module. In *Proceedings of the European conference on computer vision (ECCV)*. 3–19.

[168] Zhihao Xia, Patrick Sullivan, and Ayan Chakrabarti. 2020. Generating and exploiting probabilistic monocular depth estimates. In *IEEE conference on computer vision and pattern recognition*. 65–74.

[169] Jianxiong Xiao, Andrew Owens, and Antonio Torralba. 2013. Sun3d: A database of big spaces reconstructed using sfm and object labels. In *Proceedings of the IEEE international conference on computer vision*. 1625–1632.

[170] Ruan Xiaogang, Yan Wenjing, Huang Jing, Guo Peiyuan, and Guo Wei. 2020. Monocular depth estimation based on deep learning: A survey. In *2020 Chinese Automation Congress (CAC)*. IEEE, 2436–2440.

[171] Junyuan Xie, Ross Girshick, and Ali Farhadi. 2016. Deep3d: Fully automatic 2d-to-3d video conversion with deep convolutional neural networks. In *European conference on computer vision*. Springer, 842–857.

[172] Dan Xu, Elisa Ricci, Wanli Ouyang, Xiaogang Wang, and Nicu Sebe. 2017. Multi-scale continuous crfs as sequential deep networks for monocular depth estimation. In *IEEE conference on computer vision and pattern recognition*. 5354–5362.

[173] Dan Xu, Elisa Ricci, Wanli Ouyang, Xiaogang Wang, and Nicu Sebe. 2019. Monocular Depth Estimation Using Multi-Scale Continuous CRFs as Sequential Deep Networks. *IEEE transactions on pattern analysis and machine intelligence* 41 (2019). Issue 6.

[174] Dan Xu, Wei Wang, Hao Tang, Hong Liu, Nicu Sebe, and Elisa Ricci. 2018. Structured attention guided convolutional neural fields for monocular depth estimation. In *IEEE conference on computer vision and pattern recognition*. 3917–3925.

[175] Jialei Xu, Yuanchao Bai, Xianming Liu, Junjun Jiang, and Xiangyang Ji. 2021. Weakly-Supervised Monocular Depth Estimationwith Resolution-Mismatched Data. *arXiv preprint arXiv:2109.11573* (2021).

[176] Xianfa Xu, Zhe Chen, and Fuliang Yin. 2021. Monocular depth estimation with multi-scale feature fusion. *IEEE Signal Processing Letters* 28 (2021), 678–682.

[177] Xianfa Xu, Zhe Chen, and Fuliang Yin. 2021. Multi-scale spatial attention-guided monocular depth estimation with semantic enhancement. *IEEE Transactions on Image Processing* 30 (2021), 8811–8822.

[178] Yufan Xu, Yan Wang, Rui Huang, Zeyu Lei, Junyao Yang, and Zijian Li. 2022. Unsupervised Learning of Depth Estimation and Camera Pose With Multi-Scale GANs. *IEEE Transactions on Intelligent Transportation Systems* 23, 10 (2022), 17039–17047.

[179] Jiaxing Yan, Hong Zhao, Penghui Bu, and YuSheng Jin. 2021. Channel-Wise Attention-Based Network for Self-Supervised Monocular Depth Estimation. In *2021 International Conference on 3D Vision (3DV)*. IEEE, 464–473.

[180] Gengshan Yang, Joshua Manela, Michael Happold, and Deva Ramanan. 2019. Hierarchical deep stereo matching on high-resolution images. In *IEEE/CVF Conference on Computer Vision and Pattern Recognition*. 5515–5524.

[181] Guorun Yang, Xiao Song, Chaoqin Huang, Zhidong Deng, Jianping Shi, and Bolei Zhou. 2019. Drivingstereo: A large-scale dataset for stereo matching in autonomous driving scenarios. In *IEEE/CVF Conference on Computer Vision and Pattern Recognition*. 899–908.

[182] Guanglei Yang, Hao Tang, Mingli Ding, Nicu Sebe, and Elisa Ricci. 2021. Transformer-based attention networks for continuous pixel-wise prediction. In *IEEE conference on computer vision and pattern recognition*. 16269–16279.

[183] Lihe Yang, Bingyi Kang, Zilong Huang, Xiaogang Xu, Jiashi Feng, and Hengshuang Zhao. 2024. Depth anything: Unleashing the power of large-scale unlabeled data. *arXiv preprint arXiv:2401.10891* (2024).

[184] Nan Yang, Rui Wang, Jorg Stuckler, and Daniel Cremers. 2018. Deep virtual stereo odometry: Leveraging deep depth prediction for monocular direct sparse odometry. In *Proceedings of the European conference on computer vision*. 817–833.

[185] Xin Yang, Qingling Chang, Xinglin Liu, Siyuan He, and Yan Cui. 2021. Monocular Depth Estimation Based on Multi-Scale Depth Map Fusion. *IEEE Access* 9 (2021), 67696–67705.

[186] Xin Yang, Hongcheng Luo, Yuhao Wu, Yang Gao, Chunyuan Liao, and Kwang-Ting Cheng. 2019. Reactive obstacle avoidance of monocular quadrotors with online adapted depth prediction network. *Neurocomputing* 325 (2019), 142–158.

[187] Xin Yang, Hongcheng Luo, Yuhao Wu, Yang Gao, Chunyuan Liao, and Kwang-Ting Cheng. 2019. Reactive obstacle avoidance of monocular quadrotors with online adapted depth prediction network. *Neurocomputing* 325 (2019), 142–158.

[188] Xiaodong Yang, Zhuang Ma, Zhiyu Ji, and Zhe Ren. 2023. Gedepth: Ground embedding for monocular depth estimation. In *Proceedings of the IEEE/CVF International Conference on Computer Vision*. 12719–12727.

[189] Xinchen Ye, Xin Fan, Mingliang Zhang, Rui Xu, and Wei Zhong. 2021. Unsupervised Monocular Depth Estimation via Recursive Stereo Distillation. *IEEE transactions on image processing* 30 (2021). https://doi.org/10.1109/TIP.2021.3072215

[190] Xinchen Ye, Mingliang Zhang, Rui Xu, Wei Zhong, Xin Fan, Zhu Liu, and Jiaao Zhang. 2019. Unsupervised Monocular depth estimation based on dual attention mechanism and depth-aware loss. In *International Conference on Multimedia and Expo (ICME)*. IEEE, 169–174.

[191] Wan Yingcai, Fang Lijing, and Zhao Qiankun. 2019. Multi-scale Deep CNN Network for Unsupervised Monocular Depth Estimation. In *IEEE Annual International Conference on CYBER Technology in Automation, Control, and Intelligent Systems*. IEEE, 469–473.

[192] Mehmet Kerim Yucel, Valia Dimaridou, Anastasios Drosou, and Albert Saa-Garriga. 2021. Real-time monocular depth estimation with sparse supervision on mobile. In *Proceedings of the IEEE/CVF Conference on Computer Vision and Pattern Recognition*. 2428–2437.

[193] Pierluigi Zama Ramirez, Matteo Poggi, Fabio Tosi, Stefano Mattoccia, and Luigi Di Stefano. 2018. Geometry meets semantics for semi-supervised monocular depth estimation. In *Asian Conference on Computer Vision*. Springer, 298–313.

[194] Huangying Zhan, Ravi Garg, Chamara Saroj Weerasekera, Kejie Li, Harsh Agarwal, and Ian M. Reid. 2018. Unsupervised Learning of Monocular Depth Estimation and Visual Odometry with Deep Feature Reconstruction. *Proceedings of the IEEE computer society conference on computer vision and pattern recognition*. https://doi.org/10.1109/CVPR.2018.00043

[195] Haokui Zhang, Chunhua Shen, Ying Li, Yuanzhouhan Cao, Yu Liu, and Youliang Yan. 2019. Exploiting temporal consistency for real-time video depth estimation. In *IEEE conference on computer vision and pattern recognition*. 1725–1734.

[196] Mingliang Zhang, Xinchen Ye, and Xin Fan. 2020. Unsupervised detail-preserving network for high quality monocular depth estimation. *Neurocomputing* 404 (2020), 1–13.

[197] Ning Zhang, Francesco Nex, George Vosselman, and Norman Kerle. 2023. Lite-mono: A lightweight cnn and transformer architecture for self-supervised monocular depth estimation. In *Proceedings of the IEEE/CVF Conference on Computer Vision and Pattern Recognition*. 18537–18546.

[198] Zhoutong Zhang, Forrester Cole, Richard Tucker, William T Freeman, and Tali Dekel. 2021. Consistent depth of moving objects in video. *ACM Transactions on Graphics (TOG)* 40, 4 (2021), 1–12.

[199] Zhenyu Zhang, Zhen Cui, Chunyan Xu, Zequn Jie, Xiang Li, and Jian Yang. 2018. Joint task-recursive learning for semantic segmentation and depth estimation. In *Proceedings of the European conference on computer vision*. 235–251.

[200] Zhenyu Zhang, Stephane Lathuiliere, Elisa Ricci, Nicu Sebe, Yan Yan, and Jian Yang. 2020. Online depth learning against forgetting in monocular videos. In *IEEE/CVF Conference on Computer Vision and Pattern Recognition*. 4494–4503.

[201] Ziyu Zhang, Alexander G Schwing, Sanja Fidler, and Raquel Urtasun. 2015. Monocular object instance segmentation and depth ordering with cnns. In *Proceedings of the IEEE international conference on computer vision*. 2614–2622.

[202] Chaoqiang Zhao, Qiyu Sun, Chongzhen Zhang, Yang Tang, and Feng Qian. 2020. Monocular depth estimation based on deep learning: An overview. *Science China Technological Sciences* 63, 9 (2020), 1612–1627.

[203] Chaoqiang Zhao, Gary G Yen, Qiyu Sun, Chongzhen Zhang, and Yang Tang. 2020. Masked GAN for unsupervised depth and pose prediction with scale consistency. *IEEE Transactions on Neural Networks and Learning Systems* 32, 12 (2020), 5392–5403.

[204] Shanshan Zhao, Huan Fu, Mingming Gong, and Dacheng Tao. 2019. Geometry-aware symmetric domain adaptation for monocular depth estimation. In *IEEE conference on computer vision and pattern recognition*. 9788–9798.

[205] Shiyu Zhao, Lin Zhang, Ying Shen, Shengjie Zhao, and Huijuan Zhang. 2019. Super-resolution for monocular depth estimation with multi-scale sub-pixel convolutions and a smoothness constraint. *IEEE Access* 7 (2019), 16323–16335.

[206] Yunhan Zhao, Shu Kong, Daeyun Shin, and Charless Fowlkes. 2020. Domain decluttering: Simplifying images to mitigate synthetic-real domain shift and improve depth estimation. In *Proceedings of the Conference on Computer Vision and Pattern Recognition*. 3330–3340.

[207] Chuanxia Zheng, Tat-Jen Cham, and Jianfei Cai. 2018. T2net: Synthetic-to-realistic translation for solving single-image depth estimation tasks. In *Proceedings of the European conference on computer vision (ECCV)*. 767–783.

[208] Tinghui Zhou, Matthew Brown, Noah Snavely, and David G Lowe. 2017. Unsupervised learning of depth and ego-motion from video. In *IEEE conference on computer vision and pattern recognition*. 1851–1858.

[209] Yakun Zhou, Jinting Luo, Musen Hu, Tingyong Wu, Jinkuan Zhu, Xingzhong Xiong, and Jienan Chen. 2022. Learning Depth Estimation From Memory Infusing Monocular Cues: A Generalization Prediction Approach. *IEEE Access* 10 (2022), 21359–21369.

[210] Zhongkai Zhou, Xinnan Fan, Pengfei Shi, and Yuanxue Xin. 2021. R-MSFM: Recurrent Multi-Scale Feature Modulation for Monocular Depth Estimating. In *IEEE conference on computer vision and pattern recognition*. 12777–12786.

[211] Wei Zhuo, Mathieu Salzmann, Xuming He, and Miaomiao Liu. 2015. Indoor scene structure analysis for single image depth estimation. In *IEEE conference on computer vision and pattern recognition*. 614–622.

[212] Daniel Zoran, Phillip Isola, Dilip Krishnan, and William T Freeman. 2015. Learning ordinal relationships for mid-level vision. In *Proceedings of the IEEE international conference on computer vision*. 388–396.

[213] Yuliang Zou, Zelun Luo, and Jia-Bin Huang. 2018. Df-net: Unsupervised joint learning of depth and flow using cross-task consistency. In *Proceedings of the European conference on computer vision (ECCV)*. 36–53.

# Appendix A: Supplementary Materials (e-Pub only)

## A DEPTH SENSING TECHNIQUES USED IN DATASETS

A depth dataset contains RGB-D data (the RGB images and their corresponding ground-truth depth data). Following are the two active depth sensing techniques to acquire ground-truth depth data for real world scenes.

- **Time of Flight (ToF)**: ToF measures the time taken for a light wave to travel from the sensor to the object and back. ToF sends a pulse to measure depth. However, interference from other waves in the natural setting can result in noisy output. Otherwise, the output is more accurate than that of LiDAR.
- **Light Detection Range Sensor (LiDAR)**: This optical sensing technique measures depth using the time taken to reflect the light from the object's surface. Using a laser or LED, the sensor emits infrared light to the object's surface. LiDAR sensors have an extensive range of sensing compared to ToF. LiDAR takes less time to calculate the distance. Different settings allow multiple pulses and different wavelengths depending on the application.

Two common drawbacks of the techniques mentioned above are the limited range and less accurate results on peculiar objects with shiny or transparent surfaces [24]. Another factor is that recent cameras capture RGB images from passive sensors, while active sensors capture the corresponding depth maps. Hence, the spatial resolution of RGB and depth maps do not always match. RGB images are often available in high resolution, while depth maps are in low resolution [175]. This issue makes the training process difficult.

## B GENERIC SELF-SUPERVISED TRAINING PROCESS

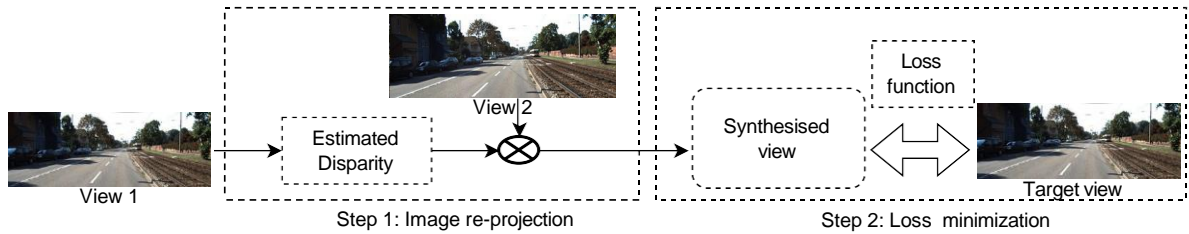

Fig. 10. Self-supervised generic training procedure.

## C SUMMARY OF THE METHODS INCLUDED FOR THE SURVEY

Table 2. Summary of selected research work for monocular depth estimation. The abbreviations follow the taxonomy presented in the paper.

| Ref. | Year | Runtime input | Training input | Archit. | Atten. | Segme. | Adapta. | Multi-Scale | Superv. | Dataset | Key Contribution | Classification based on the taxonomy |
|---|---|---|---|---|---|---|---|---|---|---|---|---|
| [55] | 2024 | Mono images | Mono images | ResNet-CLIP | N | N | N | N | 3D | KITTI, NYU-V2 | Contrastive language image pre-training for few-shot learning of depth. | TMI; RMI; FCN; 3DS; SV. |
| [136] | 2024 | Mono images | Mono images | U-Net | Y | N | N | N | 3D and self | KITTI, Flying-Things, Kubric, Autoflow, TartanAir | Using diffusion models for depth and optical-flow estimation. | TMV; RMI; FCN; (3DS, SLS); (MV,SV). |
| [140] | 2024 | Mono images | Mono images | ConvGatedRU | N | N | N | N | 3D | KITTI, NYU-V2, SUN-RGBD | Classification-regression using elastic bins. | TMI; RMI; CL; 3DS; SV. |
| [149] | 2024 | Mono images | Mono videos | ResNet | N | Y | N | N | Weakly | KITTI, DDAD, TUM, NYU-V2 | self-supervised training for dynamic scenes using pre-trained pseudo depth estimation network. | TMV; RMI; FCN; WS; SV. |
| [47] | 2023 | Mono images | Mono images | latent-based | Y | N | N | N | 3D | KITTI, DDAD, NuScenes | Zero-shot learning with latent representations for geometric embeddings. | TMI; RMI; 3DS; SV. |
| [197] | 2023 | Mono videos | Mono videos | CNN-ViT | Y | N | N | Y | Self | KITTI, Make3D | A light-weight architecture to extract multi-scale local features and attention based long-range features. | TMV; RMI; VC; SLS; SV. |
| [115] | 2023 | Mono image | Mono image | CNN | Y | N | N | Y | 3D | KITTI, NYU-V2 Argoverse, DDAD | Incorporating implicitly learned high-level patterns(discretization) within input for accurate depth. | TMI; RMI; FCN; 3DS; SV. |
| [40] | 2023 | Mono video | Mono video | CNN U-Net | N | N | N | N | 3D and Self | NuScenes, Oxford robot car | depth under challenging conditions with adversarial training. | TMV; RMV; FCN; (3DS,SLS); MV. |
| [8] | 2022 | Mono video | Mono image | CNN-ViT | Y | Patch-based | N | N | Self | KITTI, SUN3D, RGBD, MVS, Scenes11, Oxford robot car | CNN-transformer hybrid network. | TMI; RMV; VC; SLS; SV. |
| [7] | 2022 | Mono images | Mono images | CNN-ViT | Y | N | Y | N | Self | KITTI | Evaluated the generalization ability of self-supervised methods using adaptive feature fusion. | TMI; RMI; VC; SLS; SV. |
| [52] | 2022 | Mono images | Mono images | CNN | N | N | N | Y | Self | KITTI, Make3D, NYU-V2 | Data augmentation and multi-scale feature fusion method to estimate scale invariant scene depth. | TMV; RMI; FCN; SLS; SV. |
| [109] | 2022 | Mono images , GT depth | Mono images | CNN | Y | N | N | N | 3D | KITTI NYU V2 | Monocular depth estimation guided with geometric structure during supervision. | TMI; RMI; FCN; 3DS; SV. |
| [133] | 2022 | Mono images , GT depth | Mono images , GT depth | CNN | N | N | N | Y | 3D | Make3D, KITTI, NYU V2 | Multi scale feature fusion based monocular depth estimation. | TMI; RMI; FCN; FCN; 3DS; SV. |
| [10] | 2022 | Mono image | Stereo videos | CNN U-Net | N | N | | Y | Semi | Cityscapes, KITTI | Semi-supervised learning with mutual distillation. | TSV; RMI; FCN; SMS; SV. |
| [58] | 2022 | Mono images | Mono images | RNN | N | N | Y | N | Self | Cityscapes, KITTI | Recurrent modulation unit and residual upsampling for single image depth. | TMV. RMI; RWOA; OL; FL.. SV. |
| [81] | 2022 | Mono images, GT depth | Mono images, GT depth | CNN - FPN | N | N | N | Y | 3D | NYU V2, KITTI | Edge convolution for learning structural information related to depth. | TMI; RMI; FCN; SV. |
| [209] | 2022 | Mono images, GT depth | Mono images, GT depth | CNN-LSTM | N | N | Y | N | 3D | KITTI | Use of monocular cues by extracting and storing object feature vectors as a memory. | TMI; RMI; 3DS; CL. SV. |
| [155] | 2022 | Mono images, GT depth | Mono images, GT depth | U-Net | N | N | N | N | 3D | CroMo | Use of multi-modal data during 3D supervision. | TMI; RMI; FCN; 3DS. |
| [178] | 2022 | Mono videos | Mono videos | GAN | N | N | N | Y | Self | KITTI | Multi-scale gans for self supervised monocular depth estimation. | TMV; RMI; VG; SLS; SV. |
| [165] | 2021 | Mono images | Mono images | CNN | N | N | N | N | Self | KITTI, Cityscapes | Hybrid of self-supervised reprojection and multi-view cost volume matching for training. | TMV; FCN; SLS; SV. |
| [175] | 2021 | Mono images, GT depth | Mono images, GT depth | CNN | N | N | N | Y | Weakly | KITTI, NYU V2 | Training based on resolution mismatched supervision. | TMI; RMI; FCN; WS; SV. |
| [179] | 2021 | Mono videos | Mono videos | CNN | Y | N | N | Y | Self | KITTI, Make3D | Use of channel-wsie attention during self-supervised training. | TMV; RMI FCN; SLS; SV. |
| [185] | 2021 | Mono images, GT depth | Mono images, GT depth | CNN | N | N | N | Y | 3D | NYU V2 | Adaptive fusion of multi-scale depth maps. | TMI; RMI; FCN; 3DS; SV. |
| [176] | 2021 | Mono images, GT depth | Mono images, GT depth | CNN | Y | N | N | Y | 3D | KITTI, NYU V2 | Attention based multi scale feature fusion pyramid network. | TMI; RMI; FCN; DS; |
| [210] | 2021 | Mono videos | Mon videos | CNN-GRU | N | N | N | Y | Self | KITTI | Estimating depth while maintaining semantic rich feature fusion. | TMV; RMI; FCN; SLS; SV. |
| [177] | 2021 | Mono images | Stereo images | CNN | Y | N | N | Y | Self | KITTI, Make3D | Multi-scale spatial attention. | TSI; RMI; FCN; SLS; SV. |

Continued on next page

Deep Learning-based Depth Estimation Methods from Monocular Image and Videos • 3

| Ref. | Year | Runtime input | Training input | Archit. | Atten. | Segme. | Adapta. | Multi-Scale | Superv. | Datasets | Key Contribution | Classification based on the taxonomy |
|---|---|---|---|---|---|---|---|---|---|---|---|---|
| [157] | 2021 | Mono images | Mono videos | CNN | N | Y | Y | N | Self | Oxford robot car, KITTI | Joint training depth and ego-motion with scale recovery loss. | TMV; RMI; FCN; OL; |
| [189] | 2021 | Mono images | Stereo pairs | CNN | N | N | N | Y | Self | KITTI, Cityscapes, Make3D | Recursive stereo distillation via a mono network and a stereo network. | TSI; RMI; FCN; SLS; SV. |
| [161] | 2021 | Mono images | Mono videos with labeled data | CNN | Y | Y | Y | N | Self | SYNTHIA-to-Cityscapes | Domain knowledge transferring from source to target domain for adaptation. | TMV; RMI; FCN; SLS; SV. |
| [13] | 2021 | Mono image | Mono images | CNN-ViT | Y | N | | N | 3D | NYU V2, KITTI | Global processing with adaptive depth bins. | TMI; RMI; VC; 3DS; SV. |
| [16] | 2021 | Mono Video | Mono Video | ViT | Y | N | | N | 3D | ScanNet, RGBD | Multi-view feature fusion in temporal domain. | TMV; RMV 3DS; MV. |
| [26] | 2021 | Mono images | Mono images | ResNet | Y | Y | | N | Weakly | NYU V2, KITTI | Attention-based context aggregation, a classification model, soft inference to generate fine-grained depth. | TMI; RMI; AED; WS; SV. |
| [28] | 2021 | Mono images | | ViT | Y | | | Y | | | A transformer-based network that uses hierarchical representation. | TMI; RMI; ViT; MI. |
| [29] | 2021 | Mono images | Stereo image pairs | CNN U-Net | N | Y | | N | Semi | DIML/CVL, KITTI | Confidence guided network with student-teacher strategy, DIML/CVL dataset. | TSI; RMI; FCN; SMS; SV. |
| [41] | 2021 | Mono video | Mono Video | RNN | N | Y | N | Y | 3D | EventScape, MVSEC | A framework that processes multi-modal input asynchronously using RNN. | TMV; RMV; FL. |
| [25] | 2021 | Mono images | Synthetic data with GT | CNN | Y | N | N | N | Semi | Synthetic, PBN, KITTI | A framework that learns a structured representation to generalize depth estimation. | RMI; FCN; SMS; |
| [60] | 2021 | Mono videos | Mono videos | CNN | N | N | N | N | Semi | TikTok | Reconstructing human body in motion using monocular depth estimation. | TMV; RMV; FCN; SMS; MV. |
| [93] | 2021 | Stereo images | Stereo images | Cycle GAN | N | N | Y | N | Self | Oxford robot car | Adaptation for day and night monocular depth estimation. | TSI; RMI; CG; SLS; DT; SV. |
| [106] | 2021 | Mono images | Mono images | CNN | N | N | Y | Y | 3D | Middlebury, IBMS | Content adaptive multi-resolution merging. | TMI; RMI; FCN; 3DS; MI. |
| [92] | 2021 | Mono images | Mono images, GT depth | Pyramid CNN | Y | N | N | Y | 3D | NYU V2 | Multi scale attention based context aggregation module for CNNS. | TMI; RMI; FCN; 3DS; |
| [97] | 2021 | Mono images | Mono images , GT depth | CNN | N | N | Y | N | 3D | their own data | Online adaptation framework using visual SLAM. | TMI; RMI; AED; OL; MI. |
| [11] | 2021 | Mono images | Stereo images | CNN | Y | N | N | N | Self | KITTI | Two stage training strategy with ambiguity using coarse depth priors. | TSI; RMI; FCN; SV. |
| [145] | 2021 | Mono images | Mono videos, sparse GT depth | CNN-UNet | N | N | N | Y | Self | KITTI | Multi-level feature fusion with dual attention based network. | TMV; RMI; UN; SLS; SV. |
| [113] | 2021 | Mono images | Stereo images | CNN | N | N | N | Y | Self | KITTI | Self-distillation in monocular depth estimation without using an additional network. | TSI; RMI; FCN; SLS; SV. |
| [118] | 2021 | Panorama images | Mono images, GT depth | CNN-RNN | N | N | N | Y | 3D | BuildingParser | Image slicing based architecture for indoor panorama depth estimation. | TMI; RMI; 3DS; CR; MI. |
| [62] | 2021 | Mono images | Mono videos | CNN | Y | N | N | N | Self | EuRoC, NYU V2, 7-Scenes | A novel depth factorization module to adapt to rapid scale changes. | TMV; RMI; FCN; SLS; MV. |
| [72] | 2021 | Mono video | Mono video | CNN | N | Y | N | N | Self | MPI Sintel | Geometrically consistent dense depth reconstruction. | TMV; RMV; FCN; SLS; Synthetic. |
| [75] | 2021 | Mono video | Mono video | CNN U-Net | N | N | Y | N | Self | KITTI | Adaptive depth and ego-motion estimation technique for on the fly learning. | TMV; RMV; OL; SV. |
| [82] | 2021 | Mono images | Stereo image pairs | FCN | Y | N | N | N | Self | KITTI | Optimising UNet architecture with attention. | TSI; RMI; AED; SLS; SV. |
| [45] | 2020 | Mono images | Mono images | CNN U-Net | Y | N | | N | Weakly | NYU V2 HC Depth | Spatial attention block-based network archi, edge-aware loss term. | TMI; RMI; FCN; AED; WS; SV. |
| [45] | 2020 | Mono images | Mono images | FCN | N | N | Y | N | Self | Cityscapes, KITTI, NuScenes | A CNN architecture for high resolution self supervision. | TMV; RMI; FCN; SLS; SV. |
| [46] | 2020 | Mono Video | Mono Video | FCN | N | N | N | N | Semi | FlyingThings3D, Cityscapes, KITTI | A supervised loss, that minimizes reprojected distances in image space. | TMV; RMI; FCN; SMS; (SV,Synthetic). |
| [59] | 2020 | Mono Images | Mono Images | CNN | Y | N | Y | N | 3D | NYU V2, ScanNet | | TMI; RMI; FCN; SV. |
| [112] | 2020 | Mono videos | Mono videos | ConvLSTM | N | N | Y | N | Self | KITTI | Recurrent network to leverage spatiotemporal information. | TMV; OL; CL. MV. |

Table 2 – continued from previous page

| Ref. | Year | Runtime input | Training input | Archit. | Atten. | Segme. | Adapta. | Multi-Scale | Superv. | Datasets | Key Contribution | Classification based on the taxonomy |
|---|---|---|---|---|---|---|---|---|---|---|---|---|
| [95] | 2020 | Mono images | Mono images, GT depth | CNN | Y | N | N | Y | 3D | KITTI, NYU V2 | A framework that use attention-based fusion for multi-scale feature maps. | TMI; RMI; FCN; AED; 3DS; SV. |
| [148] | 2020 | Mono images | Mono images , GT depth | CNN | Y | N | N | Y | 3D | NYU V2, KITTI | Fusing multi-scale context information using information exchang network. | TMI; RMI; FCN; SV. |
| [103] | 2020 | Monocular videos | Monocular videos, GT depth | ConvGated RU | Y | N | Y | N | 3D | KITTI | Recurrent network with gated recurrent units and ConvL-STM for integrating temporal information. | TMV; RMI; OL; CL. SV. |
| [128] | 2020 | Mono images | Mono videos, GT depth | U-Net | N | N | N | Y | Joint 3D, self, weakly | | Improved depth estimation with concurrent learning using supervised, unsupervised and weakly supervised techniques. | TMV; RMI; FCN; WS. |
| [65] | 2020 | Mono images | Stereo image pairs | CNN U-net | Y | Y | N | N | Self | KITTI | Self attention and discrete disparity volume for monocular depth. | TSI; RMI; FCN; WS; SV. |
| [66] | 2020 | Mono image | Mono images | Conditional GAN | | N | N | Y | 3D | RGBD | GAN for single image depth. | TMI; RMI; CG; 3DS; MV. |
| [196] | 2020 | Mono images | Stereo images | CNN | N | N | N | Y | Self | KITTI, Cityscapes, Make3D | Rectangle convolution capturing global dependencies to preserve details in depth map. | TSI; RMI; FCN; SLS; |
| [203] | 2020 | Mono images | Stereo images | GAN | N | N | N | Y | Self | KITTI, Make3D | GAN network to elimanate occlusion using masking. | TSI; RMI; VG; SLS; |
| [206] | 2020 | Mono images | Mono and synthetic images | CNN | Y | N | Y | N | 3D | NYU V2, SUNCG | Real to synthetic domain adaptation with domain decluttering. | TMI; RMI; FCN; 3DS; DT; (MV,Synthetic) |
| [200] | 2020 | Mono images | Mono videos | CNN | Y | N | Y | N | Self | KITTI, Cityscapes | Online learning based depth estimation with an adapter to handle domain shift problem. | TMV; FCN; OL; FT; |
| [194] | 2020 | Mono images | Stereo images | CNN | N | N | N | Y | Self | KITTI, NYU V2 | Depth and visual odometry learning in self-supervised setting using stereo image pairs. | TSI; RMI; FCN; |
| [168] | 2020 | Mono images | - | | N | N | N | N | 3D | NYU V2 | To output a task agnostic probability distribution for single image depth. | TMI; RMI; 3DS; |
| [159] | 2020 | Mono images | Mono videos | CNN | N | Y | N | N | Self | KITTI | Joint self-supervision for depth, optical flow and pose. | TMV; RMI; FCN; SLS; |
| [160] | 2020 | Mono images | Mono images, semantic labels | FCN | N | Y | N | N | 3D | NYU V2, Cityscapes | Depth aggregation guided through semantic segmentation. | TMI; RMI; FCN; 3DS; |
| [3] | 2019 | Mono image | Stereo image pairs | CNN U-Net | N | N | | N | Semi | Cityscapes, KITTI | A left-right consistency based loss term for semi supervision. | TSI; RMI; FCN; SMS; |
| [20] | 2019 | Mono image | Mono video | FCN | N | Y | | N | Self | Cityscapes, KITTI | Utilising 3D geometry structure and semantics. | TMV; RMI; FCN; OL; |
| [21] | 2019 | Mono image | Mono video | CNN - DispNet | N | Y | | N | Self | Cityscapes, KITTI | Integrating both semantic and geometric information for self-supervision. | TSV; RMI; FCN; SLS; |
| [44] | 2019 | Mono images | Mono training | CNN U-Net | N | N | N | Y | Self | KITTI, Make3D | Novel appearance matching loss, auto masking technique for multi-scale monocular supervision. | TMV; RMI; FCN; SLS; |
| [48] | 2019 | Mono images | Mono images | Cycle GAN | N | N | Y | N | 3D | UnityDepth, KITTI | Leveraging large scales of synthetic images with GAN based translation. | TMI; RMI; CG; 3DS; |
| [57] | 2019 | Mono images | sparse depth maps, Mono images | CNN | N | N | N | Y | Weakly | KITTI, NYU V2 | Depth completion network using sparcity invariant operations. | TMI; RMI; FCN; WS; |
| [54] | 2019 | Mono images | Mono images , GT depth | CNN | N | N | N | N | 3D | NYU V2 | Masking input to understand how CNNS learn to monocular estimate depth. | TMI; RMI; FCN; |
| [37] | 2019 | Mono images | Stereo images | CNN | N | N | N | N | Self | KITTI | Filter pruning with binary masks to obtain light-weight CNN models for monocular depth estimation. | TSI; RMI; FCN; SLS; |
| [129] | 2019 | Mono images | Stereo images | CNN | N | N | N | Y | Self | KITTI | Enabling two networks to estimate left and right disparities for self-supervision. | TSI; RMI; FCN; SLS; |
| [122] | 2019 | Mono images | Stereo images | GAN-CRF | N | N | N | N | Self | KITTI, Cityscapes | Learning to estimate disparity maps using a CRF coupled dual GAN network. | TSI; RMI; GC; SLS; |
| [144] | 2019 | Mono images | Mono images, GT depth | CNN-CRF | N | Y | N | Y | 3D | | Contextual convolution neural network with CRF model for scene aware depth estimation. | TMI; RMI; CC; 3DS; |
| [131] | 2019 | Mono images | Stereo images | | N | N | Y | N | Self | KITTI, Cityscapes | Replacing camera pose estimation for self superivision using SLAM algorithm. | TSI; RMI; SLS; FT; |
| [124] | 2019 | Mono images | Mono images, GT depth | U-Net | Y | N | N | N | 3D | Synthetic, NYU V2 | Edge and occlusion aware depth estimation with attention based loss. | TMI; RMI; FCN; AED; 3DS; |

Continued on next page

| Ref. | Year | Runtime input | Training input | Archit. | Atten. | Segme. | Adapta. | Multi-Scale | Superv. | Datasets | Key Contribution | Classification based on the taxonomy |
|---|---|---|---|---|---|---|---|---|---|---|---|---|
| [116] | 2019 | Mono images | Stereo videos | CNN | N | N | N | Y | Self | KITTI | Sub-pixel convolution for recovering super resolution depth. | TSV; RMI; FCN; SLS; |
| [63] | 2019 | Mono images | Mono images | Conditional GAN | N | N | Y | N | Semi | NYU V2, KITTI | Semi-supervised framework in adversarial learning paradigm. | TMI; RMI; CG; |
| [79] | 2019 | Mono images | Mono images, GT depth | CNN | N | N | N | Y | 3D | NYU V2, KITTI | A network architecture that utilizes local planer guidance layers for decoding. | TMI; RMI; FCN; 3DS; SMS; |
| [152] | 2019 | Mono images | Mono images, GT depth | CNN | N | N | N | N | Semi | NYU V2, ImageNet | Semi-supervised learning method based on confidence learning. | TMI; RMI; FCN; SMS; |
| [153] | 2019 | Mono images | Stereo images | CNN | N | N | N | Y | Self | KITTI, Cityscapes | Enhanced self-supervision using cost volume and disparity refinement. | TSI; RMI; FCN; SLS; |
| [205] | 2019 | Mono images | Mono images, GT depth | CNN | N | N | N | Y | 3D | NYU V2, Make3D | Sub-pixel convolution for super resolution depth maps. | TMI; RMI; FCN; 3DS; |
| [173] | 2019 | Mono images | Mono images, GT depth | CNN-CRF | N | N | N | Y | 3D | KITTI, NYU V2, Make3D | Fusing multiple CNN outputs using conditional CRFs. | TMI; RMI; CC; 3DS; |
| [204] | 2019 | Mono images | Stereo images | CNN | N | N | Y | N | 3D | KITTI, Make3D | Domain adapted depth estimation with geometry awareness using synthetic images. | TSI; RMI; CyG; 3DS; SV. |
| [195] | 2019 | Mono videos | Mono videos, GT depth | GAN | N | N | Y | Y | 3D | KITTI, NYU V2 | Depth estimation using both spatial and temporal correlation features. | TMV; RMV; VG; OL; |
| [191] | 2019 | Mono images | Stereo images | CNN | N | N | N | Y | Self | KITTI | Multi-scale error loss for self-supervised learning. | TSI; RMI; FCN; SLS; |
| [153] | 2019 | Mono images | Stereo images | CNN | N | N | N | Y | Self | KITTI, Cityscapes | Monocular depth estimation using traditional stereo matching techniques. | TSI; RMI; FCN; SLS; SV. |
| [190] | 2019 | Mono images | Stereo pairs | CNN | Y | N | N | N | Self | KITTI | Dual attention to capturing both spatial and channel feature dependencies. | TSI; RMI; AED; SLS; |
| [162] | 2019 | Mono images | Multiview images | ConvLSTM | N | N | N | N | Self | KITTI | Monocular depth odometry estimation with ConvLSTM based architecture. | TMVS; RMI; SLS; CL; SI. |
| [164] | 2019 | Mono images | Stereo images | CNN | N | N | N | N | Self | KITTI | Overcoming ambiguous reprojection using monocular hints in self-supervised stereo learning. | TSI; RMI; FCN; SV. |
| [2] | 2018 | Mono image | Stereo image pairs | Vanilla GAN | | | | Y | Self | Cityscapes, KITTI | Monocular depth in GAN paradigm, reliable evaluation protocol. | TSI; RMI; VG; SLS; SI. |
| [4] | 2018 | Mono images | Mono image | Cycle GAN | N | N | | | 3D | Cityscapes, KITTI | Synthetic to real, domain adaptation via style transfer. | TMI; RMI; CG; DT; (SV,Synthetic). |
| [22] | 2018 | Mono images | Mono image | Conditional GAN | N | N | | N | 3D | NYU V2, KITTI | Context-aware depth estimation with adversarial training. | TMI; RMI; CG; 3DS; SV. |
| [31] | 2018 | Mono video | Mono video | ConvLSTM | N | N | | N | 3D | KITTI | A network to learn spatiotemporal mapping between image and depth. | TMV; RV; 3DS; CL. SV. |
| [32] | 2018 | Mono video | Mono video | GAN-CRF | N | N | | N | Semi | KITTI | Adversarial training with monocular video. | TMV; RMV; GC; SMS; SV. |
| [38] | 2018 | Mono images | Mono images | CNN | N | N | | Y | 3D | NYU V2, KITTI | Spacial increasing discretisation for relative depth. | TMI; RMI; FCN; 3DS; SV. |
| [53] | 2018 | Mono images | Mono images, GT depth | CNN-WSM | N | N | N | N | 3D | NYU V2 | Whole strip masking based CNN convolution operations for monocular depth estimation. | TMI; RMI; FCN; 3DS; SV. |
| [120] | 2018 | Mono images | Stereo videos | CNN | N | N | N | Y | Self | KITTI | Self supervision using trinocular assumption. | TSV; RMI; SLS; SV. |
| [50] | 2018 | Mono images | Mono images | FCN | Y | N | Y | Y | 3D | B3DO, NYU V2, SUN-RGBD | Novel framework with dense feature extractor and an attention mechanism. | TMI; RMI; AED; 3DS; (SV,MI). |
| [64] | 2018 | Mono images | Mono images | CNN | Y | Y | N | Y | 3D | NYU V2, KITTI | Attention driven loss with a network that propagates semantic information. | TMI; RMI; AED; 3DS; SV. |
| [70] | 2018 | Mono images | Stereo images | CNN | N | N | N | N | 3D | KITTI, NYU V2, DIML | Deep variational model which integrates both global and local predictions. | TSI; RMI; 3DS; |
| [193] | 2018 | Mono images | Stereo imsges, semantic labels | CNN | N | Y | N | N | Semi | KITTI | Using semantic label to obtain supervision signal for stereo based self-supervision. | TSI; RMI; FCN; SMS; |
| [207] | 2018 | Mono images | Synthetic image similar to synthetic | GAN | N | N | Y | N | 3D | KITTI, SUNCG | Using image translation to estimate depth from real images similar to synthetic images. | TMI; RMI; VG; 3DS; SV. |
| [199] | 2018 | Mono images | Mono images, GT depth | CNN | Y | Y | N | Y | 3D | NYU V2, Sun RGBD | Recursive learning for both depth estimation and semantic segmentation. | TMI; RMI; FCN; 3DS; MI. |

Deep Learning-based Depth Estimation Methods from Monocular Image and Videos

Table 2 – continued from previous page

| Ref. | Year | Runtime input | Training input | Archit. | Atten. | Segme. | Adapta. | Multi-Scale | Superv. | Datasets | Key Contribution | Classification based on the taxonomy |
|---|---|---|---|---|---|---|---|---|---|---|---|---|
| [213] | 2018 | Mono images | Mono videos | CNN | N | N | N | N | Self | KITTI | Joint learning of depth and optical flow. | TMV; RMI; FCN; SLS; SV. |
| [174] | 2018 | Mono images | Mono images, GT depth | CNN-CRF | Y | N | N | Y | 3D | KITTI | Attention-guided multi-scale feature fusion with CRFs. | TMI; RMI; CC; 3DS; SV. |
| [74] | 2018 | Mono images | Mono images | ResNet | N | N | Y | N | Self | NYU V2, KITTI | Unsupervised domain adaptation for pixel-wise regression. | TMI; RMI; FCN; SLS; SV; DT. |
| [80] | 2018 | Mono images | Mono images, GT depth | ResNet | N | N | N | N | 3D | NYU V2 | ResNet architecture and depth balanced euclidean distance based loss. | TMI; RMI; FCN; 3DS; SV. |
| [84] | 2018 | Mono images | Mono images, GT depth | CNN U-Net | Y | N | Y | Y | 3D | NYU V2, ScanNet | Deep attention based classification network for depth estimation. | TMI; RMI; AED; 3DS; SV. |
| [85] | 2018 | Mono images | Mono images, ordinal depth labels | ResNet | N | Y | Y | Y | 3D | Megadepth | Multi-view internet photo collection as training data using modern SFM and MVS. | MVS; RMI; FCN; 3DS; SI. |
| [19] | 2017 | Mono image | Mono images, GT depth | CNN-CRF | | | | N | 3D | NYU V2, KITTI | Discrete depth label classification. | TMV; RMI; CC; SV. |
| [102] | 2017 | Mono images | Mono videos, sparse GT depth | RNN-LSTM | N | N | Y | N | 3D | KITTI | Generalization using synthetic data and LSTM for global scale estimation. | TMV; RMI; 3DS; FL; SV. |
| [43] | 2017 | Mono images | Stereo image pairs | FCN | N | N | N | Y | Self | Cityscapes, KITTI | End-to-end unsupervised approach with a loss that enforce left-right consistency. | TSI; RMI; FCN; SLS; SV. |
| [76] | 2017 | Mono images | Stereo images, GT depth | ResNet | N | N | Y | N | Semi | KITTI | Using both supervised and unsupervised cues for depth estimation. | TSI; RMI; FCN; SMS; SV. |
| [154] | 2017 | Mono images | Mono images, GT depth, normals | CNN | N | N | N | N | 3D | SUN3D, RGBD-SLAM | Joint estimation on depth, motion, surface normals and optical flow. | TMI; RMI; FCN; 3DS; MI. |
| [172] | 2017 | Mono images | Mono images, GT depth | CNN-CRF | N | N | N | Y | 3D | NYU V2 | Multi-scale feature fusion with CRFs for CNNs. | TMI; RMI; CC; SV. |
| [171] | 2016 | Mono images | Stereo pairs from 3D movies | CNN | N | N | N | N | Self | Deep3D | Using stereo images from 3D movies for self-supervision. | TSI; RMI; FCN; SLS; MV. |
| [24] | 2016 | Mono image | Mono images | CNN | N | Y | | Y | Weakly | DIW, RGBD | Depth in the wild dataset. | TMI; RMI; FCN; WS; MI. |
| [39] | 2016 | Mono image | Stereo image pairs | CNN | N | N | | N | Self | KITTI | Coarse to fine depth with CNN. | TSI; RMI; FCN; SLS; SV. |
| [78] | 2016 | Mono images | Mono images, GT depth | FCN | N | N | N | N | 3D | NYU V2, make3d | Fully convolution architecture with novel up conv. blocks. | TMI; RMI; FCN; 3DS; SV. |
| [127] | 2016 | Mono images | Mono images, GT depth | _ | N | Y | N | N | 3D | Sintel, KITTI | Motion segmentation-based method to estimate depth for dynamic complex scenes. | TMI; RMI 3DS; SV. |
| [132] | 2016 | Mono images | Mono images, GT depth | CNN-RF | N | N | N | Y | 3D | Make3D, NYU V2 | CNN and random forest-based framework for monocular depth estimation. | TMI; RMI; CC; 3DS; MV. |
| [35] | 2015 | Mono images | Mono images | CNN | N | Y | | Y | 3D | NYU V2,RGBD | Common multi-scale network for depth, semantic and surface normal. | TMI; RMI; FCN; SV. |
| [91] | 2015 | Mono images | Mono images, GT depth | CNN-CRF | N | N | N | N | 3D | NYU V2, make3d | A framework using deep convolutional neural fields with continuous CRFs. | TMI; RMI; CC; 3DS; SV. |
| [83] | 2015 | Mono images | Mono images, GT depth | CNN-CRF | N | N | N | Y | 3D | NYU V2, make3d | Hierarchical CRFs for improving depth regression. | TMI; RMI; HCRF; 3DS; SV. |
| [212] | 2015 | Mono images, relative depth | Mono images | CNN | N | Y | N | N | Weakly | Intrinsic images in the Wild | Estimating ordinal depth in term of point pairs. | TMI; RMI; FCN; 3DS; MI. |
| [211] | 2015 | Mono images | Mono images, GT depth | CRF | N | Y | N | N | 3D | NYU V2, RMRC | A hierarchical representation by taking hints on the global structure of indoor scenes. | TMI; RMI; 3DS; SV. |
| [208] | 2015 | Mono images | Mono videos | CNN | N | N | N | Y | Self | KITTI | Self-supervision using monocular videos. | TMV; FCN; SLS; SV. |
| [201] | 2015 | Mono images | Mono images, 3D box annotations | CNN-MRF | N | Y | N | Y | 3D | KITTI | Joint learning, for instance, segmentation and depth ordering. | TMI; RMI; CC; 3DS; SV. |
| [36] | 2014 | Mono images | Mono images | CNN | N | N | N | N | 3D | NYU V2, KITTI | Scale invariant error for multi-scale depth prediction. | TMI; RMI; FCN; 3DS; SV. |
| [67] | 2014 | Mono video | Mono video | | N | N | N | N | 3D | Make3D, NYUD | a non-parametric depth transfer approach to infer temporally consistent depth | TMV; RMI; 3DS; SV. |
| [73] | 2012 | Mono image | Mono images | MRF | N | N | N | N | 3D | Google earth | depth using local gradient-based features and MRFs | TMI; RMI; MRF; 3DS; MV. |

## D SUMMARY OF EXISTING SURVEYS

Table 3. Summary of existing survey articles on depth estimation using monocular images and videos. NA: Reviewed Network Architectures, SM: Reviewed Supervision Modes.

| Survey | Year | Time range | Main focus | Papers covered | NA. | SM. | Taxonomy |
|---|---|---|---|---|---|---|---|
| Bhoi et al. [14] | 2019 | 2018-2019 | Comparing similarities and differences of 5 approaches | 5 | Y | Y | Only on supervision |
| Zhao et al. [202] | 2020 | 2014-2019 | An overview of the depth estimation based on deep learning | 110+ | N | Y | Based on supervision only |
| Xiaogang et al. [170] | 2020 | 2015-2020 | An overview on deep learning based approaches | 45 | N | Y | None |
| Swaraja et al. [150] | 2021 | 2012-2021 | Focus on training methodologies | 35 | Y | N | Based on architecture only |
| Ji et al. [62] | 2021 | 2012-2020 | Experimenting on different CNN architectures towards good practices | 40+ | Y | N | None |
| Ming et al. [107] | 2021 | 2014-2020 | Summary of deep learning approaches | 190+ | Y | Y | Only based on 6 years of work and training supervision only |
| Vyas et al. [156] | 2022 | - | Gaps in outdoor depth estimation | 60+ | Y | Y | Network architecture and supervision only |
| Bae et al. [6] | 2023 | - | a study on the generality of the network architectures. | - | Y | N | CNN and ViT networks only |
| Ours | 2024 | 2012-2024 | Critically reviewing most recent literature based on the pipeline, architectures and datasets, covering a broad taxonomy | 160+ | Y | Y | Based on input modality, architectures and training (supervision, datasets) |

## E LIMITATIONS OF EXISTING SURVEYS

We provide a brief summary of existing surveys on monocular depth estimation and their limitations. Bhoi et al. [14] present a survey on only five methods: four supervised and one unsupervised. It compares the similarities and differences between those methods. However, only considering a few papers does not provide broad knowledge on the topic. Xiaogang et al. [170] also present a survey highlighting only two supervision modes: supervised and self-supervised. Similarly, Swaraja et al. [150] present a study with 35 papers. Given the number of new papers published in the last few years, their survey only covers a limited scope. Meanwhile, Zhao et al. [202] present a survey with papers published up to 2020. The survey discusses datasets and evaluation metrics for monocular depth estimation. However, their taxonomy is limited to supervision modes used for training.

Ji et al. [62] review Convolutional Neural Network (CNN)-based monocular depth estimation techniques. They highlight the good practices for using CNN models. Their survey includes an experimental evaluation of the effects of different combinations of CNN encoder-decoder networks, training losses, and input data conditions such as resolution and training sample size. The major limitation is that their review only focuses on CNN-based encoder-decoder models with a limited number of papers. They did not cover the growing interest in hybrid architectures such as CNN-Long-Short-Term Memory (CNN-LSTM), CNN-Conditional Random Fields (CNN-CRFs), and Generative Adversarial Network-CRF (GAN-CRF). Ming et al. [107] present a survey of deep learning-based monocular depth estimation methods between 2014 and 2020. Their work summarises deep learning models, methods, datasets, evaluation metrics, and future trends. They have proposed a taxonomy based on training supervision and tasks (single and multi-task). However, they only covered CNN, RNN, and GAN-based architectures, which were popular then.

In their survey, Vyas et al. [156] focus on methods for outdoor depth estimation. Their work did not consider monocular depth estimation in indoor environments such as living rooms, office rooms, and industrial halls. The study was limited to about 60 papers in the literature. The survey does not include an explicit taxonomy and compares only the CNN and RNN-based methods and datasets specific to the outdoor domain. Several existing work [6, 9] study on neural network architectures for monocular depth estimation and their ability to generalize to unseen data. However, they mainly focus on CNN and transformer-based structures only.

## F SHORT DESCRIPTIONS OF COMMON DATASETS

### F.1 Datasets with real scenes

Outdoor datasets which are commonly used for autonomous driving applications are described below.

**KITTI** [105]: Contains complex scenes captured from outdoor rural and urban regions during day time. Stereo image pairs of resolution $1242 \times 376$, which contain up to 15 to 30 objects (pedestrians and vehicles) along with depth captured using LiDAR sensors, are available. They have calibrated cameras, a localization system, and a laser scanner to be synchronized to provide accurate ground-truth. KITTI has been used in many works as the benchmark dataset. It is mostly suitable for autonomous driving-related tasks. However, non-Lambertian surfaces such as reflecting and transparent surfaces, objects with large displacements due to speed, and sunny or cloudy lighting conditions are likely to have produced erroneous data. KITTI training set contains 61 scenes with 3712 training samples.

**Cityscapes** [30]: This is a large scale dataset covering urban. It includes outdoor scenes from 50 different cities and it is most suitable for semantic reasoning and segmentation-based depth estimation research for 3D scene understanding. The dataset includes both pixel-level annotation data and coarse annotations separately. The average human and vehicle density captured is higher compared to KITTI 2015.

**NYU V2 Depth** [141]: To develop a good dataset for the segmentation task NYU depth has been proposed. NYU V2 contains 1449 pairs of aligned image depth pairs, which cover RGB-D data of 646 diverse indoor scenes. Being designed for segmentation purposes, it has both ground-truth depth data and semantic labels. This is a useful in-depth estimation performed with respect to semantic segmentation. The presence of regions with small objects, fine details, and heavy occlusions in scenes can badly impact the accuracy of the ground-truth.

**EuRoC** [17]: This dataset is specifically designed for evaluating visual-inertial localization and 3D Reconstruction of industrial environments. The dataset accommodates 11 sets of stereo images captured from an aerial viewing device. Different conditions have been considered in coming up with a complete aerial view dataset, such as slow flights, dynamic flights, motion blur, and illumination. A distinct feature is the availability of both motion and 3D ground-truth for the airborne platform. The dataset includes raw data, spatio-temporally aligned ground-truth, and extrinsic and intrinsic camera parameters.

In addition, the most commonly used datasets with indoor scenes are:

**SUN3D** [169]: This dataset has been initially designed for place enteric scene understanding. It contains RGB-D data captured across indoor environments, covering full apartments. Data include RGB-D images of $640 \times 480$ resolution, together with segmentation labels, camera poses, and point clouds. While NYU V2 only covers part of a room, SUN3D provides depth and labels for the whole room or multiple rooms. This medium-scale dataset has 425 video sequences from 254 scenes inside 41 buildings. It is most appropriate for human-like interpretation tasks for indoor scenes.

**RGBD SLAM** [146]: This has been used as a benchmark dataset for evaluating visual SLAM and odometry systems. The dataset includes image sequences of an office environment and industrial hall with 39 image sequences with a resolution of $640 \times 480$ in two indoor settings. Dataset is good for indoor scene understanding. However, not much attention was given to capturing data under varying illumination conditions and at different times of the day.

## F.2 Synthetic datasets

We also highlight three synthetic datasets which contributed to breakthroughs in the field of monocular depth estimation.

**Scenes11**: A synthetic dataset which consists of virtual scenes and random geometry with motions. Being a synthetic dataset it has a more accurate ground-truth and poses. The disadvantage is the lack of realistic features [154].

**Blendswap**: A dataset developed based on blender-powered 3D objects and scenes. This covers 150 scenes from Blendsawp.com. This is comparably very small but brings a variety of indoor data ranging from cartoon-like to more real-like scenes [154].

**Sintel** [18]: An optical flow dataset on synthetic data, which includes data with long sequences of images, large motions, reflections, motion blur, defocus blur, and atmospheric effects. This was initially implemented for research on optical flow evaluation. Information extracted from Sintel, a 3D animated movie, was used to render under different rendering settings and to emulate different conditions closer to real data.

Although synthetic datasets do not always match with the laws of physics, and the light interactions, shape, and shading are not realistic, these are often used as a baseline for comparing the performance of novel methods with the existing ones.

## G SUMMARY OF EVALUATION METRICS

Table 4. Evaluation metrics used in deep learning-based methods. $N$ is the total number of pixels of test images with depth values, $d_i^*$ and $d_i$ are the ground-truth depth and estimated depth at pixel $i$ and, $t$ is the threshold.

| Metric | Metric Type | Function | Interpretation |
|---|---|---|---|
| Relative Accuracy Threshold ($\delta$ <threshold(t)) | Depth accuracy | $max(\frac{d_i}{d^*}, \frac{d^*}{d_i}) = \delta < t$ | $\delta$ is % of $d_i$ and t $\epsilon$ (1.25, $1.25^2$, $1.25^3$).Higher is better. |
| Linear Root Mean Squared Error (RMSE) | Depth error | $\sqrt{\frac{1}{|N|}\sum_{i\epsilon N}||d_i - d^*||^2}$ | Root value of the average squared error. Lower is better. |
| Log Root Mean Squared Error (RMSE Log) | Depth error | $\sqrt{\frac{1}{|N|}\sum_{i\epsilon N}||log(d_i) - log(d^*)||^2}$ | Root value of the average squared log error. Lower is better. |
| Absolute Relative Error (Abs. Rel) | Depth error | $\frac{1}{|N|}\sum_{i\epsilon N}\frac{|d_i - d_i^*|}{d_i^*}$ | Reducing the effect of large errors in the distance, where the image resolution is lower and higher magnitude of the error is more likely to occur. Lower is better. |
| Squared Relative Error (Sq. Rel) | Depth error | $\frac{1}{|N|}\sum\frac{||d_i - d_i^*||^2}{d_i^*}$ | Lower is better. |
| Mean Log Error (log10) | Depth error | $\frac{1}{|N|}\sum_{i\epsilon N}||log(d_i) - log(d^*)||$ | Mean of depth error in log. Lower is better. |